\documentclass{ieeeaccess}
\usepackage{cite}
\usepackage{amsmath,amssymb,amsfonts}
\usepackage{graphicx}
\usepackage{textcomp}

\usepackage{indentfirst}

\usepackage{algorithm}

\ifCLASSOPTIONcompsoc
\usepackage[caption=false, font=normalsize, labelfont=sf, textfont=sf]{subfig}
\else
\usepackage[caption=false, font=footnotesize]{subfig}

\usepackage{epsfig} 
\usepackage{mathptmx} 
\usepackage{pifont}

\usepackage{verbatim} 

\usepackage{algpseudocode}
\usepackage{rotating}
\usepackage{adjustbox}
\usepackage{array}
\usepackage{multirow}
\usepackage{multicol}
\usepackage{makecell}
\usepackage{threeparttable}  
\usepackage{booktabs}
\usepackage{multirow}
\usepackage{bm}
\usepackage{tabularx}

\def\BibTeX{{\rm B\kern-.05em{\sc i\kern-.025em b}\kern-.08em
    T\kern-.1667em\lower.7ex\hbox{E}\kern-.125emX}}

\begin{document}
\history{Date of publication xxxx 00, 0000, date of current version xxxx 00, 0000.}
\doi{10.1109/ACCESS.2017.DOI}

\title{Hybrid tracker based optimal path tracking system for complex road environments for autonomous driving}
\author{\uppercase{Eunbin Seo\authorrefmark{1}, Seunggi Lee\authorrefmark{1}, Gwanjun Shin\authorrefmark{1}, Hoyeong Yeo\authorrefmark{1}},
\uppercase{Yongseob Lim\authorrefmark{2}, and Gyeungho Choi\authorrefmark{3}}}
\address[1]{Daegu Gyeongbuk Institute of Science and Technology, Daegu 42988, South Korea (e-mail: \{seb5428, seunggi\_lee, shinkansan, hoyeong23\}@dgist.ac.kr)}
\address[2]{Department of Robotics Engineering, Daegu Gyeongbuk Institute of Science and Technology, Daegu 42988, South Korea}
\address[3]{Department of Interdisciplinary Engineering, Daegu Gyeognbuk Institute of Science and Technology, Daegu 42988, South Korea}
\tfootnote{This research was conducted by the DGIST institution-specific project supported by the Ministry of Science and ICT (21-BRP-09 \& 21-BRP-08).}

\markboth
{E. Seo \headeretal: Hybrid tracker based optimal path tracking system for complex road environments for autonomous driving}
{E. Seo \headeretal: Hybrid tracker based optimal path tracking system for complex road environments for autonomous driving}

\corresp{Corresponding authors: Yongseob Lim (yslim73@dgist.ac.kr) and Gyeungho Choi (ghchoi@dgist.ac.kr)}

\begin{abstract}
Path tracking system plays a key technology in autonomous driving. The system should be driven accurately along the lane and be careful not to cause any inconvenience to passengers. To address such tasks, this paper proposes hybrid tracker based optimal path tracking system. By applying a deep learning based lane detection algorithm and a designated fast lane fitting algorithm, this paper developed a lane processing algorithm that shows a match rate with actual lanes with minimal computational cost. In addition, three modified path tracking algorithms were designed using the GPS based path or the vision based path. In the driving system, a match rate for the correct ideal path does not necessarily represent driving stability. This paper proposes hybrid tracker based optimal path tracking system by applying the concept of an observer that selects the optimal tracker appropriately in complex road environments. The driving stability has been studied in complex road environments such as straight road with multiple 3-way junctions, roundabouts, intersections, and tunnels. Consequently, the proposed system experimentally showed the high performance with consistent driving comfort by maintaining the vehicle within the lanes accurately even in the presence of high complexity of road conditions. Code will be available in \underline{https://github.com/DGIST-ARTIV}.
\end{abstract}


\titlepgskip=-15pt

\maketitle
\section{Introduction}
\label{sec:introduction}
In recent years, the requirements for the driving system of the autonomous vehicle are increasing in difficulty with respect to levels of driving automation proposed by SAE \cite{sae2014taxonomy}. In level 4-5 of vehicle autonomy, the vehicle should perform the main driving and request humans to transfer control only in special circumstances or in areas where autonomous driving is not possible. For this reason, combining the two fields of perception and control algorithm is emerging as a very important area in autonomous driving. Hence, end-to-end deep learning tracking algorithms \cite{tran2019robust, neven2018towards, xing2018advances} or methods using reinforcement learning \cite{khan2019latent} have been studied, but these methods are difficult to response flexible changes in the surrounding environment.

The deep learning based lane detection and the GPS based driving path tracking were also applied to deal with complex road environments. The control method used geometric model-based path tracker, Pure pursuit and Stanley controller, which are widely used in the control field.This research proposes a path tracking system that is able to pass through not only highways and motorway roads, but also urban and complex road environments.
In high-speed and low-speed, tunnel, and steep curve, the system quickly changed the appropriate tracking method and reflected it in driving stability for different speeds. In lane detection, the algorithm designed to process the deep learning method has been applied the three lane fitting methods in parallel to provide the most optimized path.

In the paper, the modified path tracking algorithm is designed to simultaneously process vision data to GPS data, in other words, local and global coordinates. Vision and GPS based driving guidance lines cause uncomfortable driving textures when driving along the lines. Accurately following the lane is also an act that does not take into account driving stability or driver acceptance. To solve this problem, this paper suggested an optimal path tracking algorithm to ensure high driving stability and comfortableness in the presence of high complexity of road conditions.

This optimal path tracking algorithm appropriately selects the most stable tracker out of modified trackers, which is called \textit{Hybrid} tracker based optimal path tracking system in this paper. This system receives sensor information at the same time as each tracker calculation, estimates the driving environment, and transfers the control authority to the most suitable tracker. With parallel system to minimize the stability degradation. This method guarantees consistent driving stability even in the presence of various driving environments with different characteristics. Furthermore, it has the advantage that additional correction of the computational speed is not required on the hybrid system even if each tracker is modified. Consequently, this paper notes that there are several advantages of using the \textit{Hybrid} tracker based optimal tracking system: (a) it is capable of driving in a complex road environment with high performance of driving stability and accuracy; (b) additional correction of the computational speed is not required even if each tracker is modified; (c) the proposed algorithm enhances high usage of vision based lane following which can be widely used in real-world environments; (d) Finally, our selection system induces the improvements of driving stability and tracking performance despite of its simple implementation. To demonstrate the generality of proposed system, we installed the system in an autonomous vehicle and were confirmed its performance at the DGIST campus and high-speed driving proving ground. Consequently, the experimental results represents that our proposed system significantly improves accuracy and stability on all seven cases at a low computational cost.

This paper is structured as follows: Section \ref{section: related works} describes related works. Section \ref{subsection: image processing method for lane segmentation result} shows lane processing algorithm using result from deep learning. Section \ref{subsection: path tracking algorithm} introduces the modified path trackers and respective features.
Section \ref{subsection: path processing method} describes the coordinate system transformation and interpolation for the path from the perception part.
Section \ref{subsection: optimal path tracker selection} demonstrates path tracker selection.
Section \ref{section: experimental results} shows experimental results and discussions of path trackers and proposed system. Finally, Section \ref{section: conclusion} provides the conclusion.

\section{Related Works}\label{section: related works}
\newcommand{\etal}{\textit{et al}.}
\newcommand{\ie}{\textit{i}.\textit{e}.}
\newcommand{\eg}{\textit{e}.\textit{g}.}

\subsection{Path Tracking Algorithm}
Generating a path that the vehicle should take to its destination and following the generated path are the most essential parts of an autonomous vehicle. There are three types of path tracking algorithms, which are divided into geometric model-based, kinematic model-based, and dynamic model-based algorithm \cite{snider2009automatic}. Among these three path tracking algorithms, the simplest algorithm type is the geometric model-based path tracking algorithm such as Pure pursuit, Stanley controller, and Vector pursuit. 
 
The algorithms used in this paper are Pure pursuit and Stanley controller. Pure pursuit and Stanley controller use the same 2-dimension bicycle model regardless of vehicle type, so that the calculation is simple and easy to apply compared to other path tracking algorithms \cite{cibooglu2017hybrid}. Unfortunately, it was reported that these two algorithms work well only on low-speed and general roads \cite{kim2020comparative}. Conversely, when the curvature of the road turns to large or the speed of the vehicle becomes high, it is unable to follow the generated path. Therefore if both kinematic model and dynamic model are used, tracking performance can be improved in more diverse environments. However, the above simple algorithms (\ie, Pure pursuit and Stanley controller) were appropriately modified in order to achieve maximum efficiency with limited computer performance.
 
Pure pursuit's tracking performance depends on the look-ahead distance, which leads importantly to set an appropriate look-ahead distance. Otherwise, unlike other path tracking algorithms, Stanley controller does not use a look-ahead distance, so parameters have not to be tuned \cite{amer2018adaptive}. Comparing multiple path tracking algorithms \cite{snider2009automatic},  Pure pursuit shows robustness to low-speeds and large errors, but it does not work well at high-speeds \cite{park2014development, dominguez2016comparison}.
However, Stanley controller has a faster rate of convergence to the ideal path than the Pure pursuit, but there are cases where the steering angle sometimes diverges on a road with a large curvature. Thus, this study focuses on modifying the existing algorithms to minimize the disadvantages mentioned above and to validate the modified tracking algorithm in various situations.

\subsection{Lane Detection Networks}

Studies about end-to-end network performance for lane detection are as follows. PointLaneNet \cite{chen2019pointlanenet} redefined the lane structure and lane distance to facilitate the training of the network. PINet \cite{ko2020key} combined point estimation and point instance segmentation, but there were limitations in the presence of the local occlusions or unclear lanes. 
The lightweight model ENet-SAD\cite{hou2019learning} has been proposed by applying self attention distillation (SAD) methods to the existing ENet\cite{paszke2016enet}. SCNN \cite{pan2018SCNN} and ENet-SAD obtain diverse and rich contextual information to solve these limitations.
In the autonomous driving system, various methods are attempted to reduce the delay from sensor recognition to vehicle maneuvering as much as possible. Recently, deep learning is essentially considered in the perception field, but it has also disadvantages that it is not able to secure both processing speed (\ie, inference time) and performance (\ie, accuracy) due to the limitation of GPU hardware. However, processing performance varies depending on the size of the network parameter or the computational complexity of the layer. The recent deep learning methodologies have proposed various methods to solve these obstacles. ENet-SAD improved the processing speed by applying self-attention approach. Its processing speed  secured 10 times faster comparing with SCNN. In addition, a method of actively using conventional computer graphics technology has been proposed to improve the performance of deep learning algorithms. Thus, this paper proposed fast and accurate lane processing algorithm using the results of ENet-SAD. As a result, low-latency and highly reliable lane detection system has been designed and validated in the presence of diverse road conditions.

\subsection{Ensemble method}
Recognizing lanes in raw images and extracting information about lanes in segmentation images are necessary processes. In machine learning, ensemble method combines predictions from multiple models rather than a single model, so that the final result achieves better performance. 
There are various methods such as voting and bootstrap aggregating (Bagging) in the ensemble method.
The voting method compares the results predicted by various algorithms for the same dataset and chooses the final result. There are also two types of voting: hard and soft voting. Hard voting simply chooses the results that received the most votes among the predictions from different models. Soft voting selects the highest result by summing the probability of the predicted results from different models. Moreover, Bagging is a method of voting to determine the final results predicted by one algorithm on various datasets sampled differently by allowing overlap. In Teow's handwriting recognition research, the best results came out by adopting soft voting \cite{teow2002robust}. Soft voting increased the accuracy and stability of aggregation as a result of prediction in signal segments \cite{li2017bearing}. Thus, this paper adopted soft voting that can be parallel processing guaranteed fast computational speed in lane fitting algorithm.
\section{Approach}\label{section: approach}
\subsection{Sensor layout}
\begin{figure}[hbt]
\centering
    \includegraphics[width=8cm]{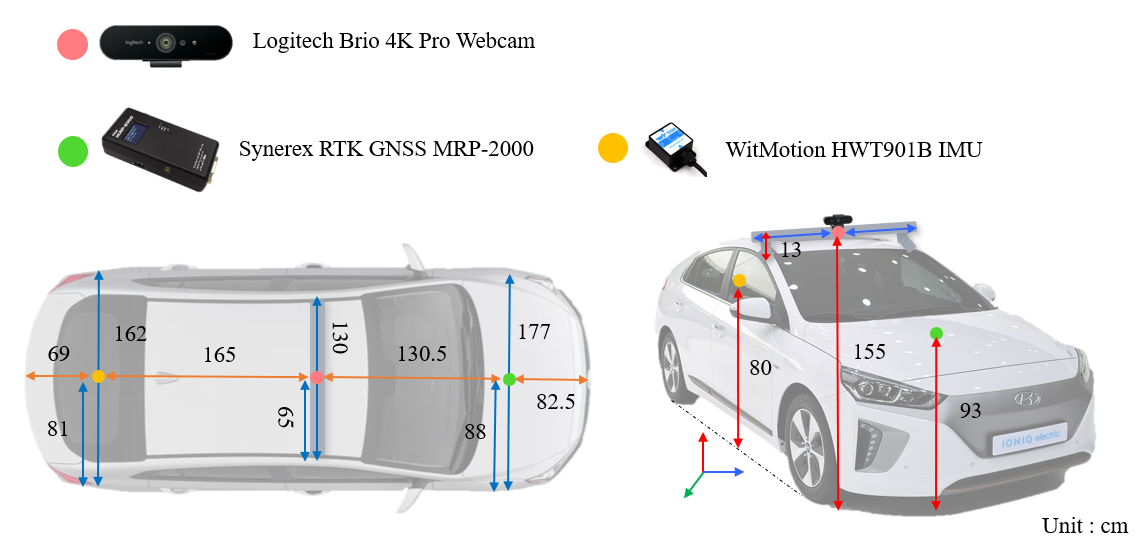}
\caption{Sensor installation on the test vehicle.}
\label{fig:sensor layout}
\end{figure}

The locations of each sensor are shown in Fig.\ref{fig:sensor layout}. Three sensors were used: camera, GPS, and IMU. Logitech's Brio 4K Pro Webcam for camera, Synerex's RTK GNSS MRP-2000 for GPS, and WitMotion's HWT901B for the IMU were utilized. 
The camera, GPS and IMU were installed at a distance of 155cm, 93cm, 80cm from the ground and 213cm, 82.5cm, 378cm from the front part, respectively shown in Fig. \ref{fig:sensor layout}.
The entire autonomous driving system of the test vehicle is equipped with one main computing unit and one discrete deep learning inference unit. Vehicle control and decision were operated by the embedded computer of Nuvo-8108GC, and a computer based on Titan Xp GPU for deep learning based lane detection algorithm was used. Detailed specifications for each sensor are listed in TABLE \ref{tab:sensor specification}.

\begin{table}[hbt]
    \begin{center}
        \footnotesize
        \caption{Specifications of sensors.}
        \renewcommand{\arraystretch}{1.5}
        \label{tab:sensor specification}
        \begin{tabular}{c|c}
        \Xhline{3\arrayrulewidth}
        Sensor & Specification \\
        \hline
        \begin{tabular}[c]{@{}c@{}}Camera\\ (Logitech Brio 4K Pro Webcam)\end{tabular} & FoV 90 degree, up to 60 frame \\
        \hline
        \begin{tabular}[c]{@{}c@{}} Real Time Kinematic GPS\\ (Synerex MRP-2000)\end{tabular} & \begin{tabular}[c]{@{}c@{}} RTK Accuracy : 0.010m +1ppm CEP\\ Fix time: 10 $\sim$ 60 sec \end{tabular} \\
        \hline
        \begin{tabular}[c]{@{}c@{}}IMU\\ (WitMotion HWT901B)\end{tabular} & 
        \begin{tabular}[c]{@{}c@{}} Angle Accuracy (after calibrated): \\ X, Y-axis: 0.05\textdegree (Static), \\ X, Y-axis: 0.1\textdegree (Dynamic) \end{tabular} \\
        \hline
        \begin{tabular}[c] {@{}c@{}} Computer\\ (Nuvo-8108GC)\end{tabular}  & \begin{tabular}[c]{@{}c@{}} CPU: Intel Xeon E 2176G, \\ GPU: NVIDIA RTX 2080 Ti, \\ 32GB RAM \end{tabular} \\
        \hline
        \begin{tabular}[c]{@{}c@{}} Computer\\ (Deep learning Inference) \end{tabular} & \begin{tabular}[c]{@{}c@{}} CPU: Intel I7-9700F, \\ GPU: NVIDIA TitanXp, \\ 32GB RAM \end{tabular} \\
        \Xhline{3\arrayrulewidth}
        \end{tabular}
    \end{center}
\end{table}

\subsection{Fast Optimal Lane Processing Algorithm}\label{subsection: image processing method for lane segmentation result}
\subsubsection{Data Preparation}
A numerous and high-quality dataset is required for deep learning network to avoid overfitting. 
First of all, ENet-SAD with the published lane dataset CULane \cite{pan2018SCNN} and TuSimple \cite{Tusimplewebsite} have been trained in this research. ENet-SAD inferred the better results when trained on CULane containing urban, rural and highway road types. CULane contains a total of 133,235 images, including training set of 88,880, validation set of 9,675 and test set of 2,782. This study developed an annotation tool that had the same file structure as CULane, and gathered approximately 45,000 images for road types in Korea.

The lane has linearity and the constant width of the lane. Since ENet-SAD learned the characteristics of the lane, the lane can be recognized even if there are objects or shadow covering the lane. The augmented data are also able to be obtained in a range that keeps the properties of lane. A total of 11 types of augmentation were applied without spatial information damage, including weather changes, brightness changes, shadow occlusion and resolution changes. As a result, total dataset consisted of 600,000 images with CULane and the Korean urban dataset. More dataset was obtained through augmentation, leading to the improved lane detection performance.

\subsubsection{Post-process for Segmentation Image}
\begin{figure}[hbt]
\centering
\subfloat[ ]{\includegraphics[width=3.5cm, height = 3cm]{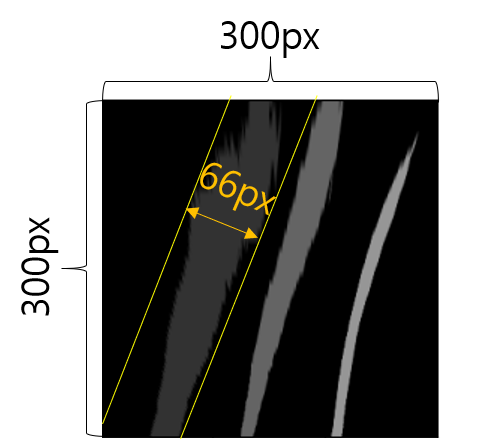}}
\subfloat[ ]{\includegraphics[width=3.5cm, height = 3cm]{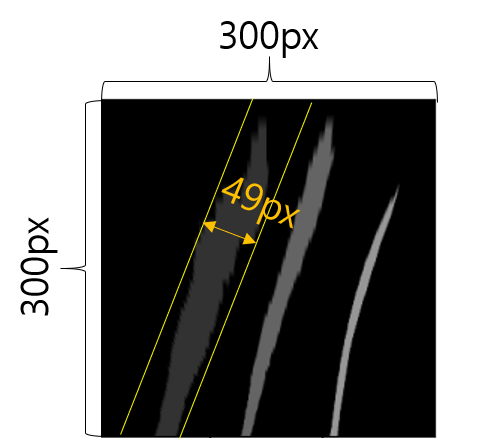}}
\caption{Before and after using erosion filter (a) segmentation image without erosion filter, (b) segmentation image with erosion filter.}
\label{fig: erosion filter comparison}
\end{figure}

The post-processing for lane segmentation images inferred from the ENet-SAD. The post-processing consists of inverse perspective mapping(IPM) and erosion filter. Due to the characteristic of camera, parallel lanes meet at the vanishing point and straight lines with the same thickness tend to become thinner as approaching the vanishing point. Images taken by camera have the distortion from the real circumstances described above and this distortion narrows the range of steering, which is able to impair steering sensitivity. To solve this phenomenon, IPM was used and it was possible to preserve the thickness and parallelism of straight lanes by converting to top view images. During converting the curve lanes, 
there is an area with thicker lanes which adversely affects the computational speed of the lane fitting process. The segmentation image with  thin lanes could be obtained using the erosion filter. Therefore, this process leads to faster computational speed in lane fitting because it reduces the number of pixels that need to be processed. Lane fitting will be detailed in section \ref{subsubsection: lane fitting}. The Fig. \ref{fig: erosion filter comparison} shows enlarged images of raw output from the ENet-SAD multiplied by a constant. The lanes are separated by different colors. Comparing the leftmost lane of Fig. \ref{fig: erosion filter comparison}(a) and Fig. \ref{fig: erosion filter comparison}(b), the thickest part is 66 pixels and 49 pixels, respectively. By simplifying data, the computational cost could be reduced.

\subsubsection{Lane Fitting}\label{subsubsection: lane fitting}

The lane fitting finds the regression function that best represents the pixel coordinates detected as lane in the segmentation image. Specifically, the methods used in lane fitting are linear, quadratic and cubic least squares fitting. Equation (\ref{equation: least-squares polynomial fitting}) expresses a least squares polynomial fitting function. The coefficient values (\ie, $c_i, i=1,2, $ \dotso, $m$) for minimizing the squared error was derived from (\ref{equation: least-squares polynomial fitting}).
\begin{align}
    \label{equation: least-squares polynomial fitting}
    \begin{pmatrix}
    p_0(x_1) & p_1(x_1) & p_2(x_1) & \dotso & p_m(x_1) \\
    p_0(x_2) & p_1(x_2) & p_2(x_2) & \dotso & p_m(x_2) \\
    \vdots & \vdots & \vdots & \ddots & \vdots \\
    p_0(x_n) & p_1(x_n) & p_2(x_n) & \dotso & p_m(x_n)
    \end{pmatrix}
    \begin{pmatrix}
    c_1 \\
    c_2 \\
    \vdots \\
    c_m
    \end{pmatrix}
    =
    \begin{pmatrix}
    y_1 \\
    y_2 \\
    \vdots \\
    y_n
    \end{pmatrix}
\end{align}
\\
where $m$ means degree, $c_i(i=1,2, $ \dotso, $m)$ is coefficients,  $p_i(x) (i=1,2,$ \dotso, $m)$ is $x$ to the $i$ power respectively.

\begin{figure}[hbt]
\centering
    \includegraphics[width=8cm]{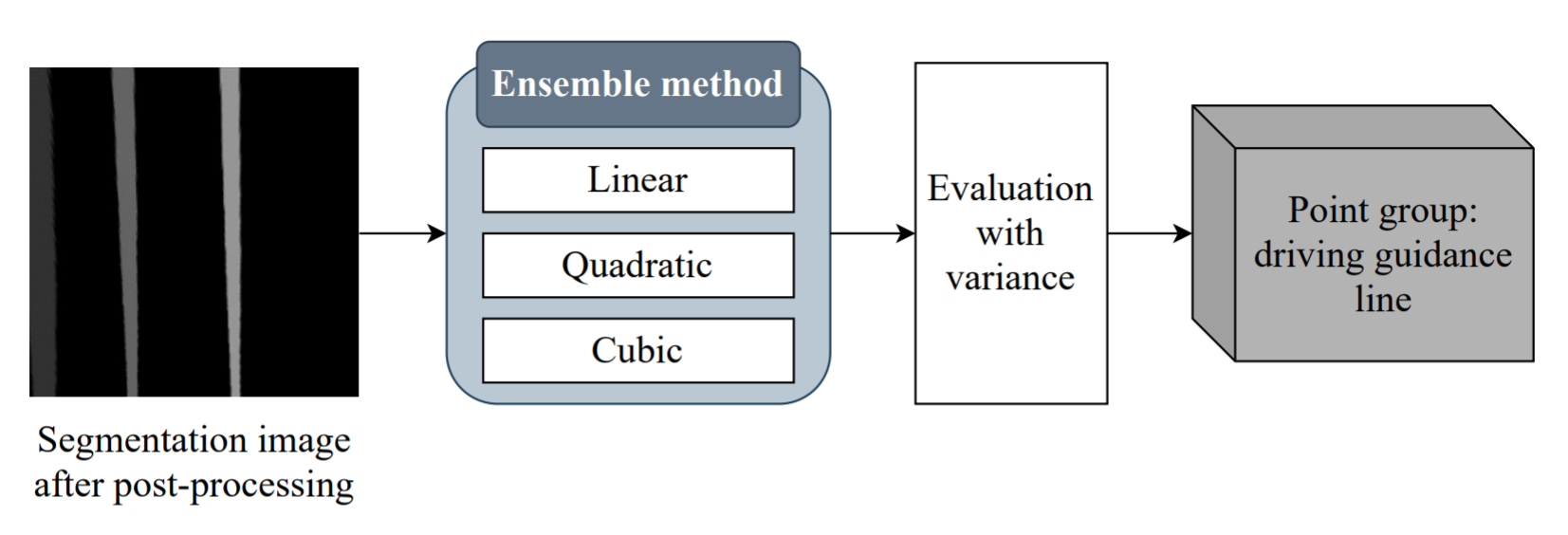}
\caption{Procedures of the lane fitting algorithm.}
\label{fig: procedure in lane fitting}
\end{figure}

The analyzed lanes consist of four lanes, which are defined as left-left lane, left lane, right lane, and right-right lane. Coefficient values were found by using linear, quadratic, and cubic least squares fitting of all four lanes. 
Some pixel coordinates were sampled by applying the coefficients obtained from the three fitting functions. Using the obtained variance as metric, the mentioned four lanes were expressed by selecting the fitting function with the smallest variance.
Based on the fitting function that best represents lanes, 31 points are obtained for each lane, and the driving guidance line is provided in the form of a point group using the fitting function corresponding to the left lane and right lane. Although one of the two main lanes is not detected, driving guidance line (hereinafter referred to as a path) can be provided by adding an offset because a lane has certain width. The procedure for vision recognition parts configured the seamless system, allowing the tracker to process the path directly. The process of overall lane fitting algorithm can be seen in Fig. \ref{fig: procedure in lane fitting}. Algorithm \ref{algorithm: lane detection} 
shows a pseudo code that contains the overall fast optimal lane processing described in the section \ref{subsection: image processing method for lane segmentation result}.
\begin{algorithm}[hbt]
	\caption{Overall fast optimal lane processing algorithm} \label{algorithm: lane detection}

	\begin{algorithmic}[1]
	
	\renewcommand{\algorithmicrequire}{\textbf{Input:}}
    \Require{$I$: Segmentation image from deep learning network}
    \renewcommand{\algorithmicrequire}{\textbf{Output:}}
    \Require{$O$: Driving guidance line consisting of points}
    \State $I_p$: Top view image after post-processing for $I$
    \State $w$: Width of $I$
    \State $h$: Height of $I$
    \State $L_{coord}$: Two-dimensional lists of coordinates for lane points
    \State $L_{func}$: Two-dimensional lists of function for lane
	\State $I_p$ $\leftarrow$ Run IPM 
    \State $I_p$ $\leftarrow$ Apply the erosion filter
		\For {$iteration=1,2,\ldots, w$}
			\For {$iteration=1,2,\ldots, h$}
				\State Get the pixel coordinates segmented by lane \State from $I_p$ and store them in the $L_{coord}$
			\EndFor
		\EndFor
		\For{$iteration(i) = 1,2,3,4$}
		    \State Run lane fitting algorithm on the $L_{coord}[i]$
		    \State Store the best function expression in the $L_{func}[i]$
		\EndFor
	    \If {left lane and right lane are exist}
	            \State $O$ $\leftarrow$ Calculate the center points with \State $L_{func}[1]$ and $L_{func}[2]$
	   \Else
	       	    \If {only left is exist}
	   	            \State $O$ $\leftarrow$ Calculate the left lane 
	   	            \State points with $L_{func}[1]$ and add offset
	   	        \Else
	   	            \If {only right is exist} 
    	   	            \State $O$ $\leftarrow$ Calculate the right lane 
    	   	            \State points with $L_{func}[2]$ and add offset
    	   	       \Else
    	   	       	     \State $O$ $\leftarrow$ Lane does not exist
	   	           \EndIf
	            \EndIf
	   \EndIf
	\end{algorithmic} 
\end{algorithm}

\subsection{Path Tracking Algorithm}\label{subsection: path tracking algorithm}
\subsubsection{Pure Pursuit}

\begin{figure}[hbt]
\centering
    \includegraphics[width=8cm, height=6cm]{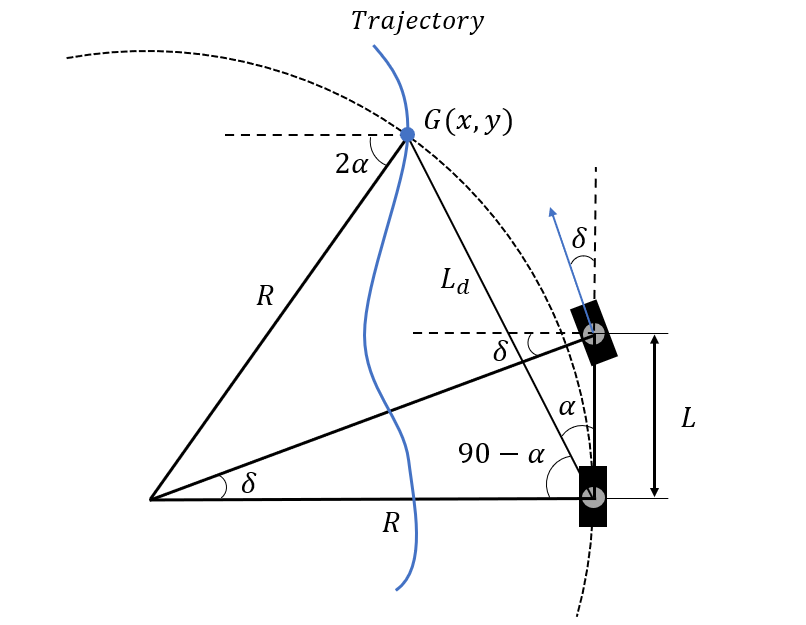}
\caption{Geometric explanation of Pure pursuit.}
\label{fig:pure pursuit geometry}
\end{figure}

Pure pursuit has been recognized one of the path tracking algorithms that follow the target point. In this algorithm, a vehicle is treated as a bicycle model, and the geometric explanation of the Pure pursuit is shown in Fig. \ref{fig:pure pursuit geometry}. 
The basic calculation method of Pure pursuit is as follows.

When drawing a circle with the radius of the look-ahead distance ($ld$) based on the rear wheel of the vehicle, the point that overlaps the given path is taken as the target point. The steering angle ($\delta$) of the vehicle is calculated using $\alpha$, which indicates the difference between the direction toward the rear wheel of the vehicle and the direction toward the look-ahead distance. At this time, the steering angle can be expressed as (\ref{equation: result delta}).

\begin{align}
    \label{equation: result delta}
    \delta = \tan^{-1}(\frac{2L\cdot sin(\alpha)}{ld})
\end{align}
\\
where, $L$ represents wheel base of the vehicle and $ld$ means the look-ahead distance respectively.

There are static and variable look-ahead distances. With a static look-ahead distance, the Pure pursuit achieves poor results because the target point is always fixed in the same radius regardless of the vehicle speed. Therefore, a variable look-ahead distance  was used in order to achieve better results and expressed as a function of velocity in this study. The variable look-ahead distance was expressed as (\ref{equation: look-ahead distance}).

\begin{align}
    \label{equation: look-ahead distance}
    ld = 190 + 100\cdot[\frac{1}{exp\{-(\frac{v-20}{15})\}}-0.5]
\end{align}
\\
Moreover, a PID controller based filter was added to prevent sudden bouncing or shaking of the steering angle. Anti-windup logic was added to prevent the divergence of error due to integral calculation. 
The compensator generating a smooth profile of the steering angle was also implemented as described in Algorithm \ref{algorithm: pure pursuit}.

\begin{algorithm}[hbt]
	\caption{Steering angle compensation algorithm} \label{algorithm: pure pursuit}

	\begin{algorithmic}[1]

	\renewcommand{\algorithmicrequire}{\textbf{Input:}}
    \Require{$\delta$: Steering angle calculated with Pure pursuit and filter}
    \renewcommand{\algorithmicrequire}{\textbf{Output:}}
    \Require{$com\_\delta$: Steering angle after compensation}
    \State $T$: Threshold
		\If {$\delta$ - $prev\_steer$ > $T$}
		    \State $com\_\delta$ = $prev\_steer$ + $T$
		\ElsIf {$prev\_steer$ - $\delta$ > $T$}
		    \State $com\_\delta$ = $prev\_steer$ - $T$
	    \Else {}
	        \State $com\_\delta$ = $\delta$
    	\EndIf
	    \State $prev\_steer$ = $com\_\delta$
	    \newline
	    \Return {$com\_\delta$}
	\end{algorithmic} 
\end{algorithm}

In this study, the modified Pure pursuit was developed to follow the vision and GPS based path provided by the perception part.
There are two types of pure pursuit depending on sensor used. Pure pursuit using a vision based path is indicated as Pure pursuit (vision), and Pure pursuit using a GPS based path is indicated as Pure pursuit (GPS).
Both Pure pursuit (vision) and Pure pursuit (GPS) become unable to drive on very large curvature roads, but stability of steering angle is highly secured on straight or slight curve. Pure pursuit (vision) is more responsive than Pure pursuit (GPS) due to fast computational speed of vision. However, Pure pursuit (vision) is unable to compute further path than Pure pursuit (GPS) which can be driven in areas without lanes.

\subsubsection{Stanley Controller \label{stanley controller}}

\begin{figure}[hbt]
\centering
    \includegraphics[width=8cm, height=6cm]{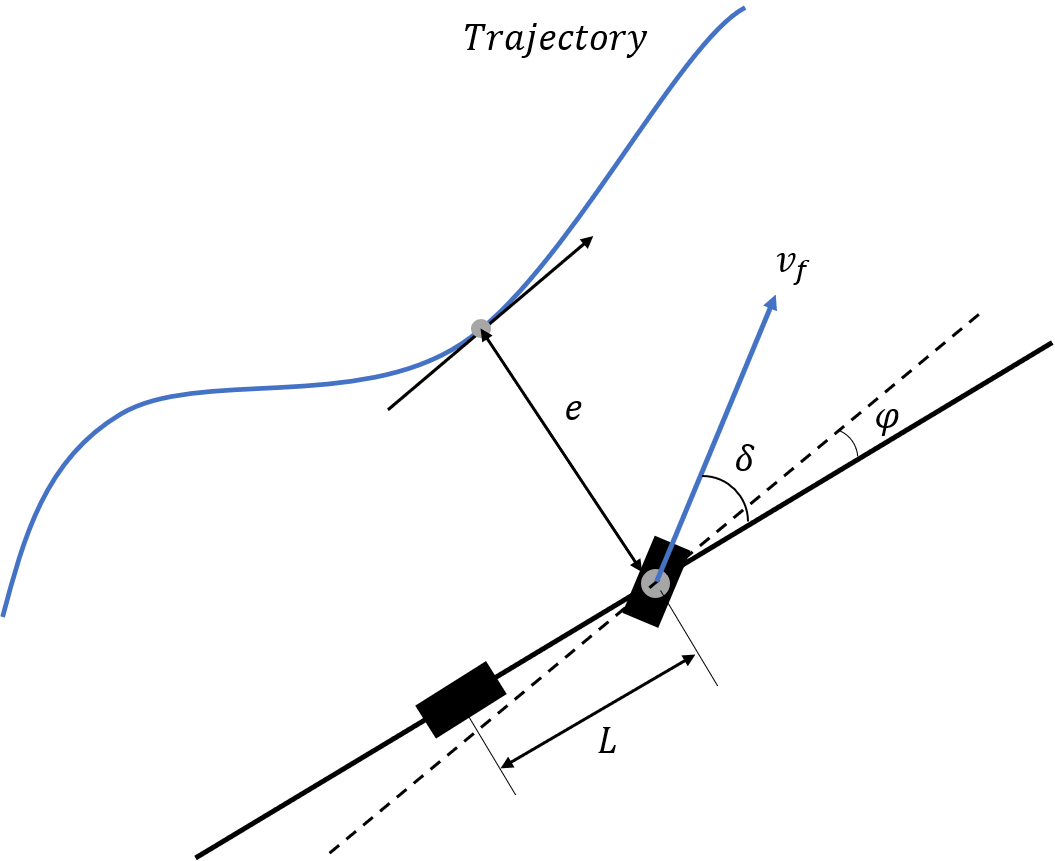}
\caption{Geometric explanation of Stanley controller.}
\label{fig:stanley controller geometry}
\end{figure}

Like the Pure pursuit, the Stanley controller calculates the steering angle based on the bicycle model. Two types of errors are used to calculate the steering angle. The two errors consist of a cross-track error and a heading error. The cross-track error refers to the minimum distance between a given path and a front wheel of the vehicle. The heading error refers to the difference between the direction vector of the path and the direction vector of the vehicle. Using these two errors, the steering angle can be obtained and a geometric explanation for this algorithm is shown in Fig. \ref{fig:stanley controller geometry}.

\begin{align}
    \label{middle_delta}
    \delta (t) = \theta_e (t) + \theta_d (t) = \varphi (t) + \tan^{-1}\frac{ke(t))}{k_s + v_f(t)}
\end{align}
\\
where $\varphi (t)$($\theta_e (t)$) is heading error, $e(t)$ is cross-track error, $v_f(t)$ is the speed of the vehicle, $k$ and $k_s$ are gain parameters respectively. $k$ and $k_s$ in parameter types are expressed in (\ref{middle_delta}). Since it is difficult to implement perfect tracking performance with only these two parameters, $k_1$ and $k_2$ should be added before $\theta_e (t)$ and $\theta_d (t)$, respectively. Therefore, the modified Stanley controller expression can be written the same as (\ref{final_delta}).

\begin{align}
    \label{final_delta}
    \delta (t) = k_1 \theta_e (t) + k_2 \theta_d (t) = k_1 \varphi (t) + k_2 \tan^{-1}\frac{ke(t))}{k_s + v_f(t)}
\end{align}
\\
This modified Stanley controller was used to follow the global path provided by the HD map. 
Stanley (GPS) and Stanley (vision) represent Stanley controller using GPS based path and vision based path respectively.
Stanley (GPS) receives the current position of the vehicle through GPS and the position of the front wheel of the vehicle can be calculated using $L$, the wheel base of the vehicle. In case of using the modified Stanley (GPS), the vehicle shakes a lot. Nevertheless, Stanley (GPS) is necessary because it can be driven in very large curvatures.

Stanley controller does not use the vision based path due to the method of determining the steering angle. As described above, Stanley controller uses the heading error and the cross-track error to calculate the steering angle. Since the vision based path starts from the front wheel of the vehicle, the cross-track error is not able to exist as zero. With Stanley (vision), the steering angle must be calculated only from the heading error. Therefore, Stanley (vision) was not used because of its low tracking performance.

\subsection{Path Processing Method}\label{subsection: path processing method}
The cameras, GPS and HD map were used to generate paths. 
GPS and HD map are essential to get the path, when lane detection was not able to properly perform or in the case of large curvature road. It was explained in the Section \ref{subsection: image processing method for lane segmentation result} that the path could be retrieved from raw image through the camera.
The process of creating a vision based path included a process in which the tracker uses the path immediately.
In order to use the GPS based path directly in the tracker like the vision based path, coordinate transformation is required.
The process of smoothly connecting paths becomes also necessary for more stable driving. Therefore, path processing method includes the coordinate transformation and interpolation.

\subsubsection{Path Processing Method based on Vision (Pure Pursuit)}
The processing method of path based on vision used in Pure pursuit  will be explained as follows. The points obtained in section \ref{subsection: image processing method for lane segmentation result} are assigned a corresponding distance estimated from the camera extrinsic parameter. The look-ahead distance varies with velocity of vehicle and the look-ahead point also changes. The target point, the closest point to the look-ahead point, affects the steering angle. Since the points are not continuous, the target point changes from time to time, which leads to instability in steering. In order to ensure that the target point is located as close to the look-ahead point as possible, the number of points was increased by selecting the two closest points to the look-ahead point and using linear interpolation between the two points. The stability of the steering angle is compensated by tracking the target point at the same location as the look-ahead point.

\subsubsection{Path Processing Method based on GPS (Stanley Controller)}
\begin{figure}[hbt]
\centering
    \includegraphics[width=8cm]{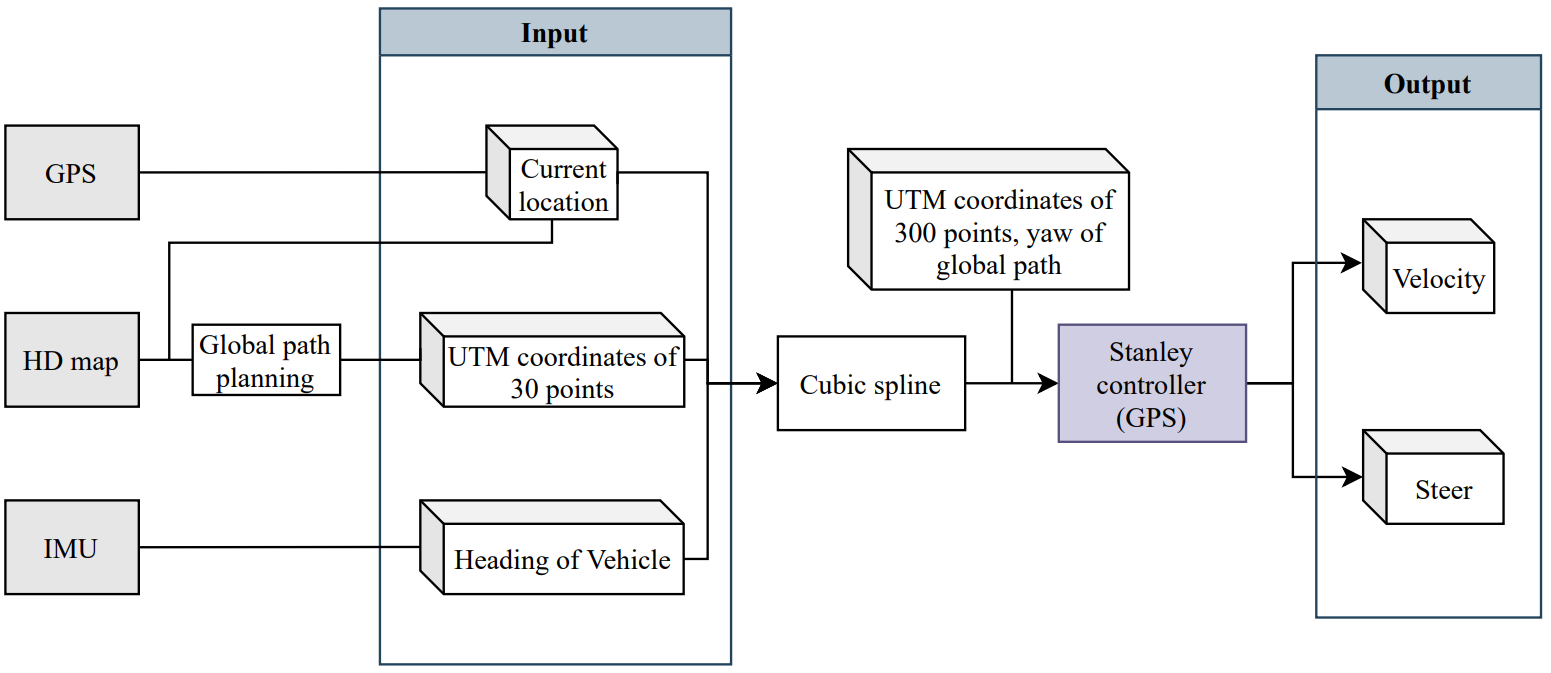}
\caption{Path processing method based on GPS used in Stanley controller.}
\label{fig: path preprocessing gps stanley}
\end{figure}

The processing method of global path used in Stanley controller is shown in Fig. \ref{fig: path preprocessing gps stanley}.
First of all, the yaw value of the vehicle is obtained using IMU. The global path is received from the HD map and provides 30 points with an interval of about 1 meter based on the current location of the vehicle. In order to use this path in Stanley controller, it is necessary to connect 30 points smoothly in a curve. Therefore, cubic spline interpolation was used to approximate these points as curves. Cubic spline interpolation represents a method of smoothly connecting given points using a cubic polynomial. By applying this, 30 points that come in at intervals of 1 meter are interpolated at intervals of about 0.1 meter, and all yaw values corresponding to each point are calculated. Using the interpolated path and vehicle's yaw value, the distance to the point on the path closest to the current vehicle and yaw at which point can be calculated. 
Using these values, the steering angle is finally obtained by the Stanley controller.

\subsubsection{Path Processing Method based on GPS (Pure Pursuit)}
\begin{figure}[hbt]
\centering
\subfloat[ ]{\includegraphics[width=3.5cm, height = 3cm]{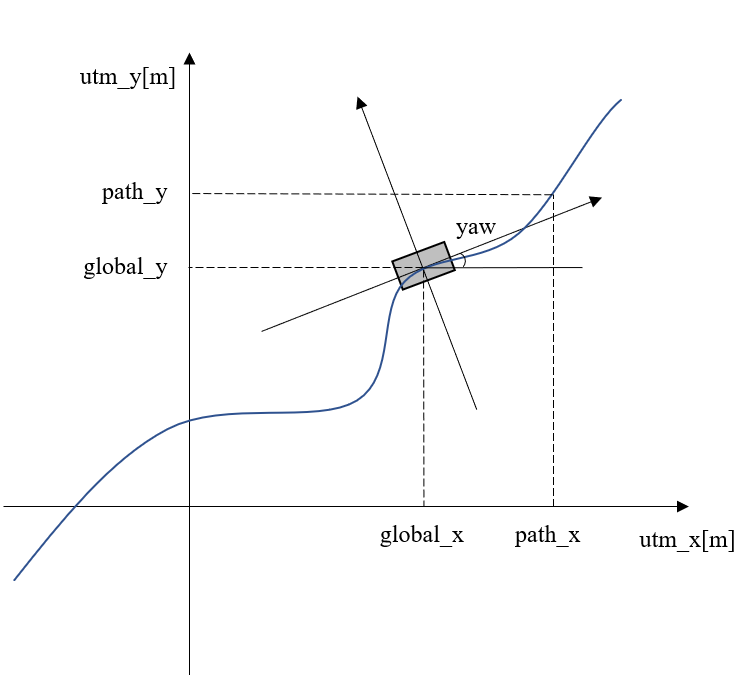}}
\subfloat[ ]{\includegraphics[width=3.5cm, height = 3cm]{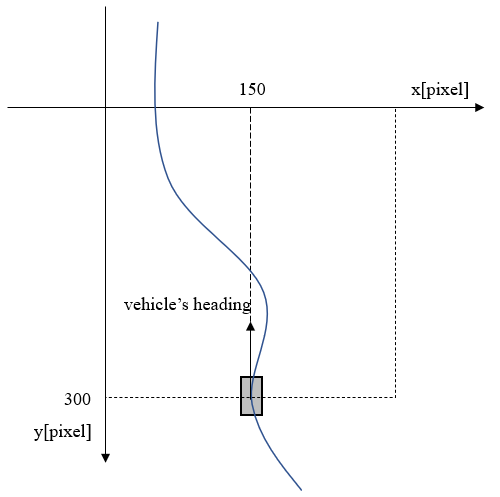}}
\caption{Path coordinate transformation (a) UTM coordinate system, (b) pixel image coordinate system.}
\label{fig: path coordinate transformation}
\end{figure}

\begin{figure}[htb]
\centering
    \includegraphics[width=8cm]{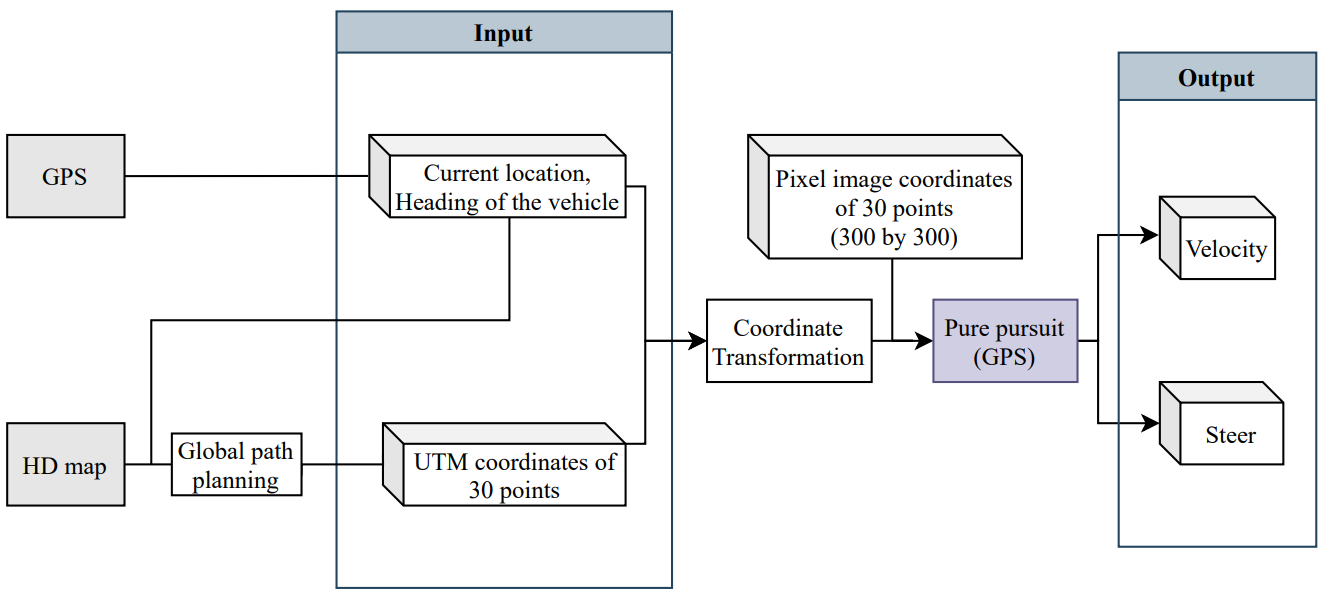}
\caption{Path processing method based on GPS used in Pure pursuit.}
\label{fig: path preprocessing gps pure}
\end{figure}

The processing method of the global path used in Pure pursuit provides 30 points at about 1 meter intervals based on the current location. The coordinate system of the global path uses the UTM coordinate system and the coordinate system used in the Pure pursuit requires local path. Therefore, the heading of the vehicle was always set to be the same as the direction of the y-axis, and coordinate transformation converting the global path into a local path was used. Since 300 by 300 pixel image coordinate was used, scaling was performed to fit the units of the coordinate system. These path coordinate transformations are shown in Fig.\ref{fig: path coordinate transformation}, and the coordinate transformation equations are formulated in (\ref{expression: coordinate transformation 1}) and (\ref{expression: coordinate transformation 2}). The data-flow diagram for the path processing method based on GPS used in Pure pursuit is shown in Fig. \ref{fig: path preprocessing gps pure}.

\begin{align}\begin{split}
    new\_path\_x = 150 - 10\cdot [cos\{-(90+yaw)\}\cdot (path\_x - \\ 
    global\_x)-sin\{-(90+yaw)\}\cdot (path\_y - global\_y)]
    \label{expression: coordinate transformation 1}
\end{split}\end{align}

\begin{align}\begin{split}
    new\_path\_y = 300 + 10\cdot [cos\{-(90+yaw)\}\cdot (path\_x - \\
    global\_x)-sin\{-(90+yaw)\}\cdot (path\_y - global\_y)]
    \label{expression: coordinate transformation 2}
\end{split}\end{align}

\subsection{Optimal Path Tracker Selection}\label{subsection: optimal path tracker selection}
The most important part of hybrid tracker based optimal path tracking system becomes optimal path tracker selection. Optimal path tracker selection chooses an appropriate path tracking algorithm among Pure pursuit (GPS), Pure pursuit (vision), and Stanley (GPS) considering instability of steering angle, GPS reliability, vision reliability, and road information. 
The selected trackers can be changed in short intervals alternately because this algorithm ensures real-time and selects the optimized tracker.
The instability of steering angle is predetermined based on the driving tendency of the Stanley controller and Pure pursuit. GPS determines the reliability of the sensor using horizontal dilution of precision (HDOP) and RTK precision information. Vision checks the information of the path received from section \ref{subsection: image processing method for lane segmentation result} and determines the detection reliability of the lane itself inside the system. It also determines road information such as curvature, roundabout, and intersection of the path to be driven based on HD map. The key role in the optimal path tracker selection is to observe the information provided by the sensors and to switch securely the path tracker. 

\begin{figure}[hbt]
\centering
\subfloat[ ]{\includegraphics[width=8cm, height = 4.0cm]{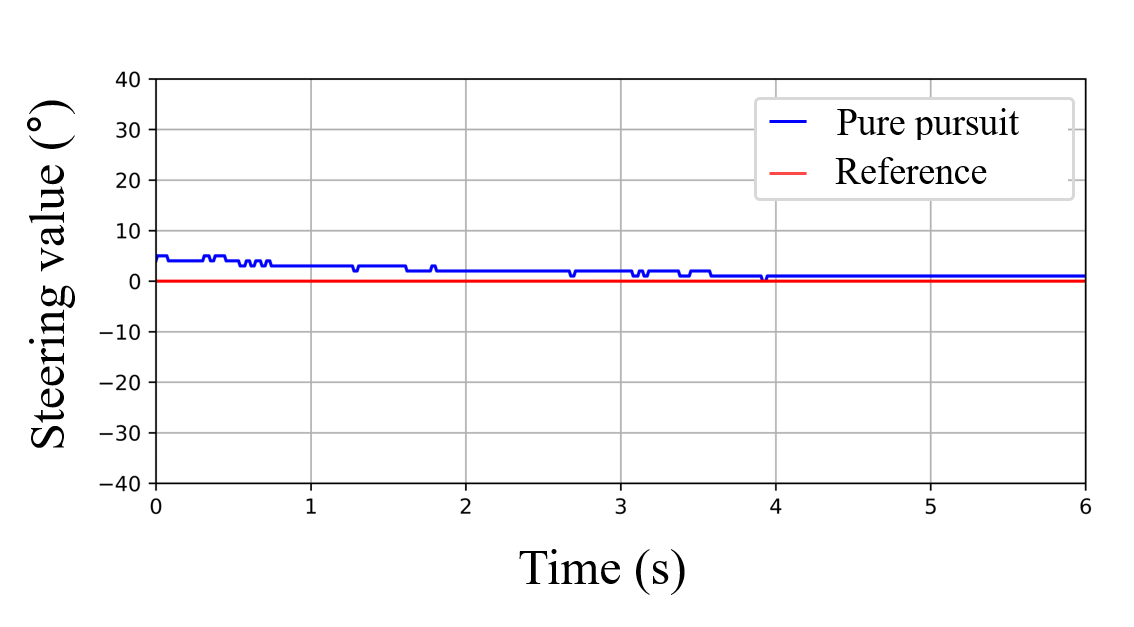}}\hfill
\subfloat[ ]{\includegraphics[width=8cm, height =4.0cm]{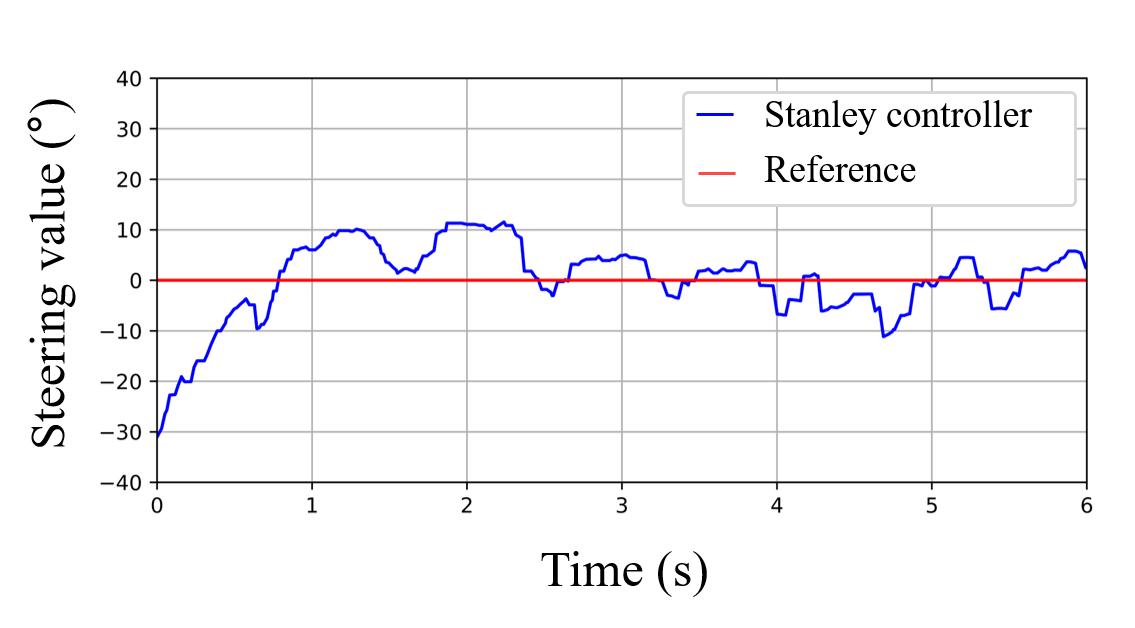}}\hfill
\caption{Steering at straight road (a) steering angle using Pure pursuit, (b) steering angle using Stanley controller.}
\label{fig: steering at straight road}
\end{figure}

The instability of the steering angle is determined according to the following results shown in Fig. \ref{fig: steering at straight road}. Comparing the steering values, the vibration of the steering angle using Pure pursuit becomes smaller than using the Stanley controller. When Pure pursuit was used, it converged to 0 degrees in 4 seconds, whereas when Stanley was used, it oscillated within 10 degrees and -10 degrees even after 4 seconds. 
It can be deduced that the driving stability would be better if the Pure pursuit with higher stability of steering angle was used in the range of driving with in-track maintained.

In this paper, GPS state information is named as GPS reliability, HD map based road information is named as HD map data and vision based lane detection state information is named as lane reliability. Each evaluating method is explained as follows.

First, HDOP refers to the degree of interfering with the positional precision of the horizontal coordinate by parsing from the information of NMEA 0183 of GPS. Therefore, HDOP and RTK precision information was obtained from GPS, and GPS reliability was evaluated by combining them. The normal operating standard of HDOP was set to 3 or less experimentally. The information of RTK precision is divided into 3 levels, which the highest level (Fixed) was treated as 2, the lower levels in turn were treated as 1 (Float) and 0 (No Fixed).
The Fixed option was chosen in this research as the highest level because GPS only works well with Fixed, which compensates for signals using LTE.
If HDOP is less than 3 and the RTK precision information satisfies Fixed option, the GPS determines only that it is operating normally as shown in TABLE \ref{tab: Truth table to determine whether GPS is available}. 

\begin{table}[hbt]
    \begin{center}
        \footnotesize
        \caption{Criteria on determination of the GPS availability.}
        \renewcommand{\arraystretch}{1.4}
        \label{tab: Truth table to determine whether GPS is available}
        \addtolength{\tabcolsep}{2pt}
        \begin{tabular}{c|c|c}
        \Xhline{3\arrayrulewidth}
        RTK precision information & HDOP & GPS reliability \\
        \hline
        \multirow{2}{*}{No Fixed(0)} & $HDOP < 3$ & \ding{53}(0) \\
        \cline{2-3} & $HDOP \geq 3$ & \ding{53}(0) \\
        \hline
        \multirow{2}{*}{Float(1)} & $HDOP < 3$ & \ding{53}(0) \\
        \cline{2-3} & $HDOP \geq 3$ & \ding{53}(0) \\
        \hline
        \multirow{2}{*}{Fixed(2)} & $HDOP < 3$ & $\bigcirc$(1) \\
        \cline{2-3} & $HDOP \geq 3$ & \ding{53}(0) \\
        \Xhline{3\arrayrulewidth}
        \end{tabular}
    \end{center}
\end{table}

The method of evaluating the lane reliability is explained as follows. In the 300 by 300 window, the location of the current vehicle was set to (150,0) and the vision based path in the direction of heading was provided as point group information. By expressing the location and point group information as an image, it was possible to determine the pixel distance as metrics for the relationship between the path and the current vehicle location. 

The pixel coordinates of the x-axis were scaled at 0.03 meter per pixel.
The width of a road equals 100 pixels. Considering width of the vehicle, if the vehicle is off one-fifth the width of the road from its current location, it could be deduced that the test vehicle invaded the center line. 
Therefore, if the Equation (\ref{expression: self assessment}) is satisfied, the vision based path is determined to be stable and 1 is provided as lane reliability.
\begin{align}
    |150-lpts| < W_r/5
    \label{expression: self assessment}
\end{align}
\\
where $lpts$ is x coordinates of lane points and $W_r$ is width of a road in the pixel coordinates respectively.

When lanes are unclear in the intersection and the roundabout, it is difficult for vision to recognize them. In tunnel or a road on a hillside, GPS status is not good, which leads GPS based path to a low reliability. Therefore, the driving stability should be improved by receiving information of road terrain in advance and selecting the optimal path tracking algorithm.

By using the optimal path tracker selection with these information, each of the disadvantages was compensated and solved simply by switching the tracking algorithm following each path of vision and GPS.


The criteria of this system for selecting the tracking algorithm using the data described above is summarized in TABLE \ref{tab:hybrid following system truth table}.

\begin{table}[hbt]
    \begin{center}
        \caption{Criteria of optimal path tracker selection.}
        \renewcommand{\arraystretch}{1.4}
        \label{tab:hybrid following system truth table}
        \resizebox{8cm}{!}{%
        \begin{tabular}{c|c|c||c}
        \Xhline{3\arrayrulewidth}
        GPS reliability & HD map data & Lane reliability & Selected path tracker \\
        \hline
        0 & * & * & \begin{tabular}[c]{@{}c@{}}Pure pursuit\\(vision)\end{tabular}\\
        \hline
        1 & 1 & 1 & \begin{tabular}[c]{@{}c@{}}Stanley controller\\(GPS)\end{tabular}\\
        \hline
        1 & 1 & 0 & \begin{tabular}[c]{@{}c@{}}Stanley controller\\(GPS)\end{tabular}\\
        \hline
        1 & 0 & 1 & \begin{tabular}[c]{@{}c@{}}Pure pursuit\\(vision, GPS)\end{tabular}\\
        \hline
        1 & 0 & 0 & \begin{tabular}[c]{@{}c@{}}Pure pursuit\\(vision, GPS)\end{tabular}\\
        \Xhline{3\arrayrulewidth}
        \end{tabular}%
        }
        \begin{tablenotes}
        \small
        \item '*' means that it doesn't matter if there is any value (0 or 1).
        \end{tablenotes}
    \end{center}
\end{table}

\begin{figure*}[htb]
\centering
    \includegraphics[width=17cm]{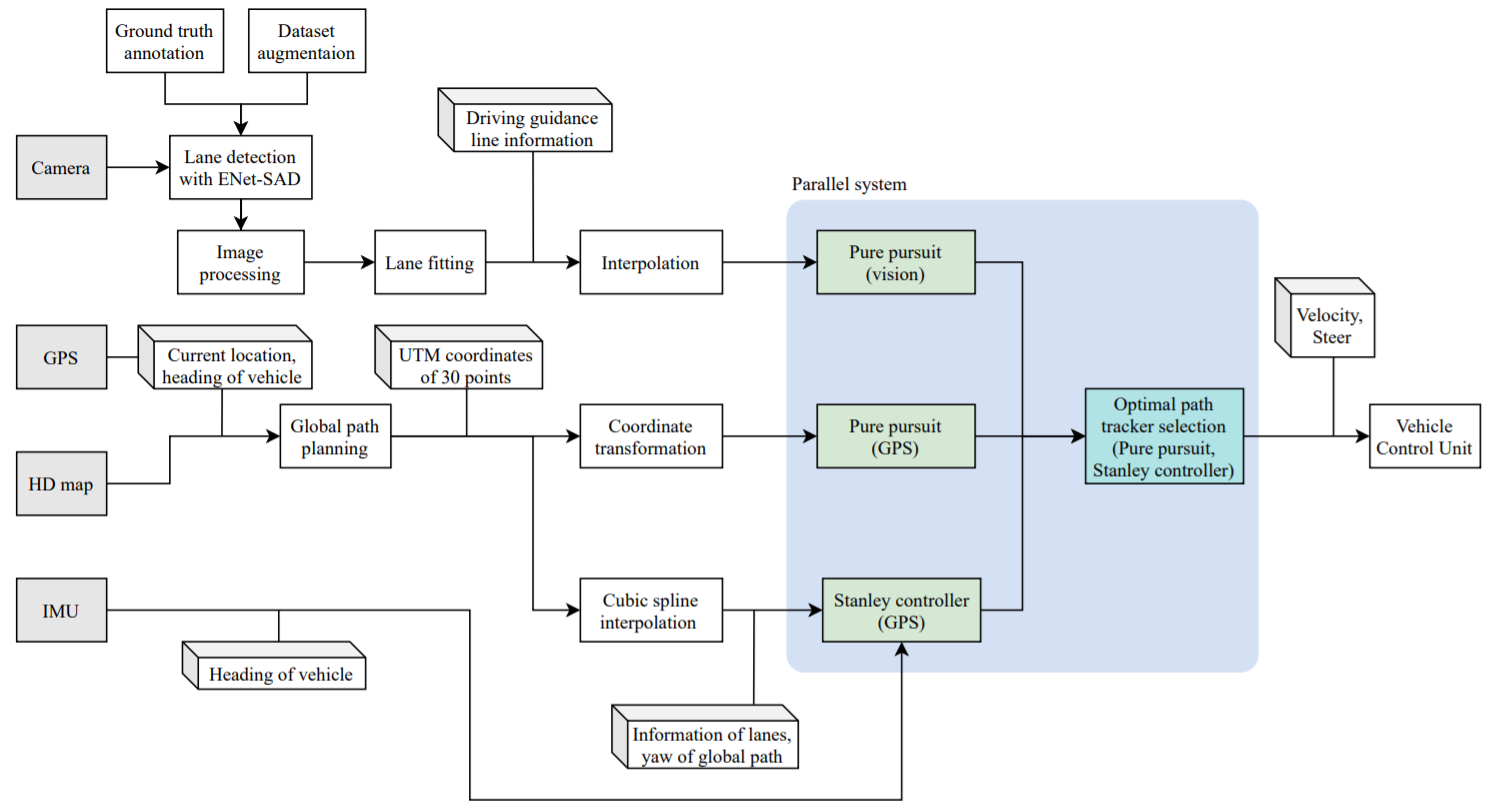}
\caption{General overview of the Hybrid tracker based optimal path tracking system.}
\label{fig:general overview of the system}
\end{figure*}


Each part from perception to control was described in detail above. In order to reduce the computational cost, the perception and control sections were closely connected, and the trackers were configured as parallel systems. An overview of the architecture of Hybrid tracker based optimal path tracking system used in this paper is shown in Fig. \ref{fig:general overview of the system}.

\section{Experimental results}\label{section: experimental results}

This section includes the performance analysis about the proposed Hybrid tracker based optimal path tracking system (hereinafter referred to as Hybrid). All of the tested algorithms were mounted on Hyundai Ioniq Electric 2020 and were tested on a shuttle route of DGIST and on high-speed circuit proving ground. Tested roads comprise a variety of roads, including two-lane or four-lane roads, intersection, as well as steep curve, tunnels, roundabout and so on. In addition, the shuttle route includes both uphill, flat and downhill, enabling acceleration testing. The high-speed circuit consists of three-lane one-way and banked curve whose radius is 100 meters. The roads in DGIST can be assumed as Korean urban road and the high-speed circuit might be assumed highways. Therefore, tested paths were considered good road conditions for verifying the system at low and high speeds (\ie, 0 km/h to 100 km/h).

The DGIST shuttle route was analyzed into normal road environments and complex road environments.
Roadways relatively easy to follow among the DGIST shuttle route are called the normal road environment. The three types of normal road environment were selected as straight, slight curve of radius 157.5 meters and steep curve of radius 53 meters because tracker is able to drive properly on these roadways using only one of the vision and GPS based path.

Sections of the road relatively difficult to follow among the DGIST shuttle route are called the complex road environment, and the four sections were designated as straight road with multiple 3-way junctions, roundabout of radius 14 meters, tunnel, and intersection. The reasons why the sections were selected as the complex road environment will be explained as follows.

Unlike roads belonging to the normal road environment, the intersection and roundabout have very large curvature and the cameras cannot recognize the lanes in the road of large curvature. In addition, since lanes occasionally disappears for straight roads with multiple 3-way junctions, GPS based path is required.
Due to the low reliability of GPS in the tunnel, the vision based path is only available.

\subsection{Performance of Fast Optimal Lane Processing Algorithm}
\begin{figure}[hbt]
\centering
\subfloat[ ]{\includegraphics[width=8cm, height = 1.8cm]{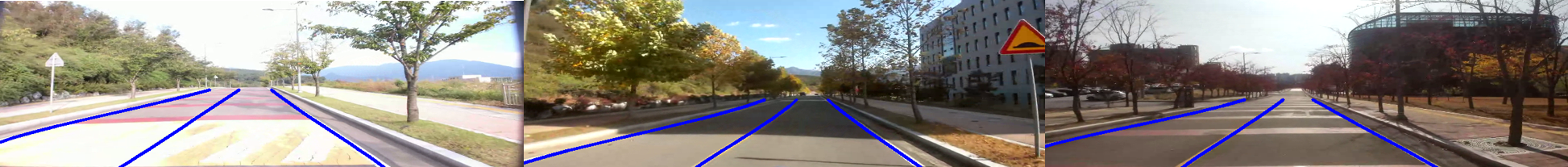}}\hfill
\subfloat[ ]{\includegraphics[width=8cm, height =1.8cm]{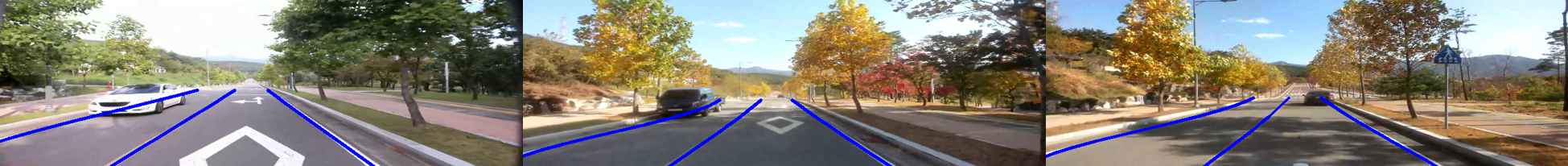}}\hfill
\subfloat[ ]{\includegraphics[width=8cm, height = 1.8cm]{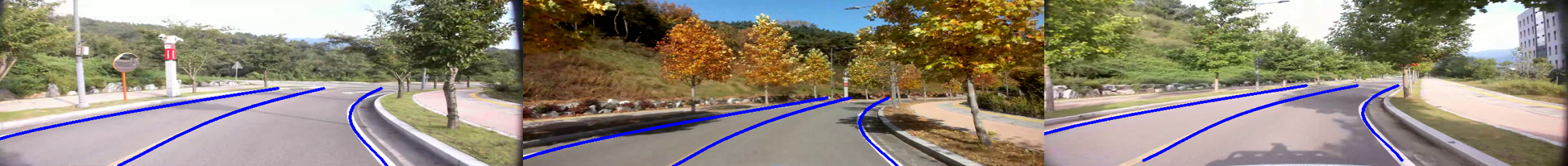}}\hfill
\subfloat[ ]{\includegraphics[width=8cm, height = 1.8cm]{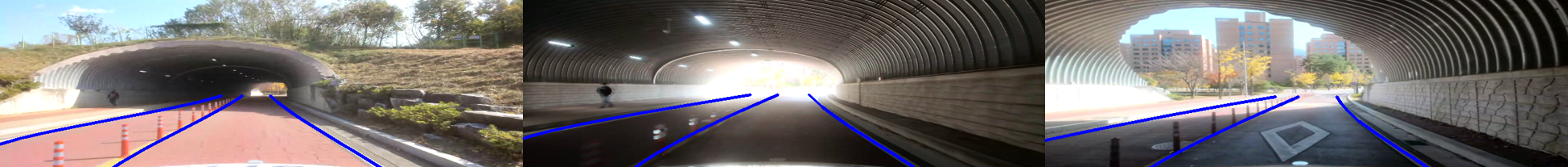}}\hfill
\caption{Estimation results after applying the fast optimal lane processing algorithm with diverse road conditions (a) straight roads, (b) roads with lane covered, (c) curve roads, (d) tunnel.}
\label{fig:fast opitmal lane processing algorithm}
\end{figure}

The performance of fast optimal lane processing algorithm will be explained in this section. The Fig. \ref{fig:fast opitmal lane processing algorithm} shows the image overlapping the raw image with the lane of top view image processed by the fast optimal lane processing algorithm. In the Fig. \ref{fig:fast opitmal lane processing algorithm}, 
(a), (b), (c) and (d) show the results of lane processing algorithm at straight road, roads with lane covered by vehicles, curve road, tunnel respectively. The results of straight roads consist of uphill, flat and downhill roads. The results of the tunnel show that the vehicle passes through the tunnel. The algorithm recognizes lanes despite of structures or vehicles on the road, such as speed bump, eye-inducing rods, vehicles. It can be deduced that this algorithm robustly recognizes lanes in various road environments and find the best fitting regression function.

The time-averaged to infer lanes from ENet-SAD is taken 23.53ms. The time-averaged to fit the pixel coordinates is measured 4.42ms. The sum of several delays is approximately found 5ms. Therefore, the final runtime is 32.95ms. 

\subsection{Evaluation on Real-world}

The modified Pure pursuit, Stanley controller and Hybrid system have been tested and analyzed in complex road conditions. Two types of driving guidance line, the vision based path and the GPS based path can be used by each path tracking algorithm. There are four possible cases; Pure pursuit (vision), Pure pursuit (GPS), Stanley (vision), and Stanley (GPS). As for these four cases, the other three cases excluding Stanley (vision) which had poor performance have been studied in this paper. The detailed reason for excluding the Stanley (vision) was explained in section \ref{stanley controller}.

Therefore, the three modified trackers were optimized for coordinate transformation and post-processing algorithms of various sensor parts. Tracking performance and driving stability have been evaluated by comparing the Hybrid and three modified trackers with ideal path based on HD map. 
$Success$ represents that the tracking algorithms drove in the designated area without hitting the curb or moving into the opposite lane

To determine the tracking performance of each tracker, RMSE ($lateral$), RMSE ($longitudinal$) and $distance$ were used in metric unit. These values represent the difference between the ideal path and the real driving path. 
RMSE ($lateral$) and RMSE ($longitudinal$) stand for the root mean square error of latitude and longitude, respectively. The "$distance$" signifies the difference between the ideal path and the real driving path.

Driving stability has been determined using the RMSE ($yaw$) and RMSE ($steer$) in degree unit. RMSE ($yaw$) expresses the difference between the yaw of the vehicle and the yaw of HD map. RMSE ($steer$) means the difference between the steering angle directly driven by a human in an ideal path and the steering angle calculated by each path tracking algorithm. 
The $yaw$ indicates vehicle's heading, and the $steer$ means a value that determines the direction of the vehicle. The small RMSE ($yaw$) value represents that the planned driving direction becomes similar to the ideal driving direction. The small RMSE ($steer$) value also implies that the steering value calculated by the tracking algorithm becomes similar to steer of the human, since it is compared to the driving smoothly of human. Therefore, if both values show small, it implies driving stability is high because the algorithm follows the path smoothly without fluctuation. Thus, the driving stability can be evaluated with values of RMSE ($yaw$) and RMSE ($steer$).

These six experimental results were summarized in TABLE \ref{tab: normal} and \ref{tab: complex} to find easily out the differences, and the Top-2 results for each column was emphasized in these tables.
\subsubsection{Test Results Under Normal Road Environment \label{subsubsection: tracking performance normal}}

\begin{figure*}[hbt]
\centering
\subfloat[ ]{\includegraphics[width=3.5cm, height = 3.5cm]{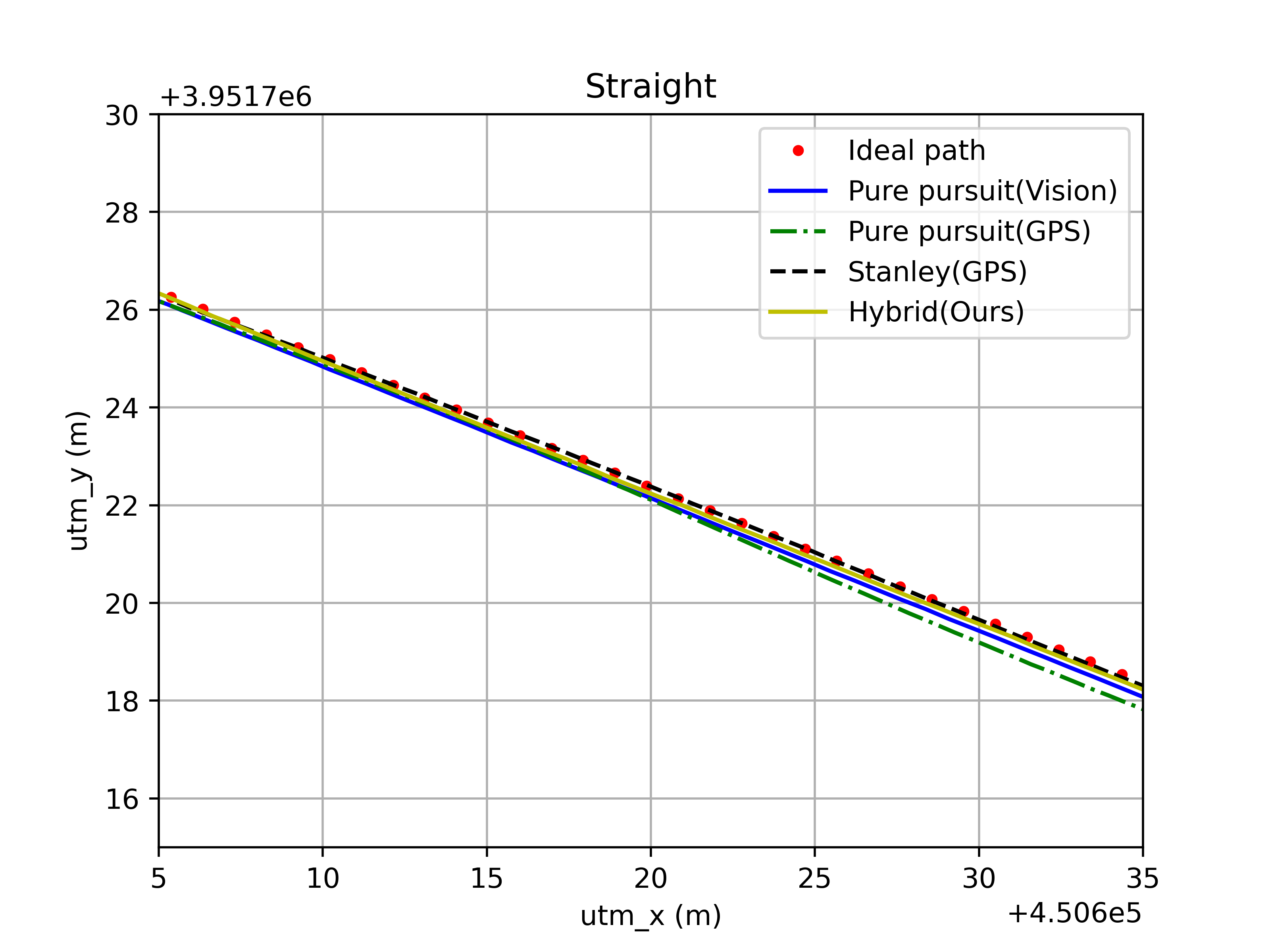}}
\subfloat[ ]{\includegraphics[width=3.5cm, height = 3.5cm]{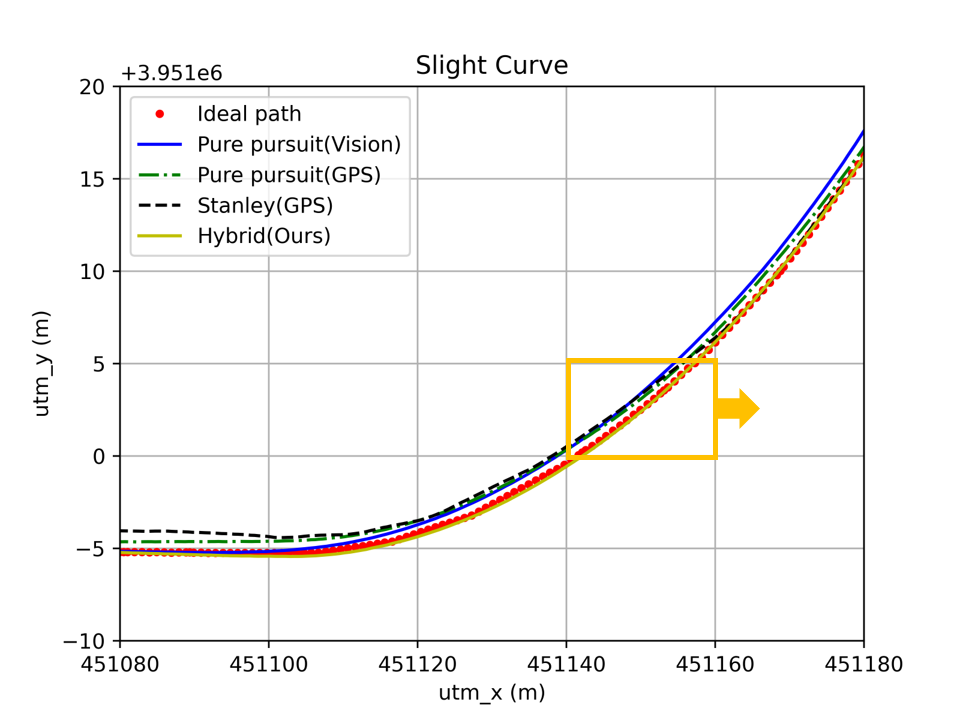}}
\subfloat[ ]{\includegraphics[width=3.5cm, height = 3.5cm]{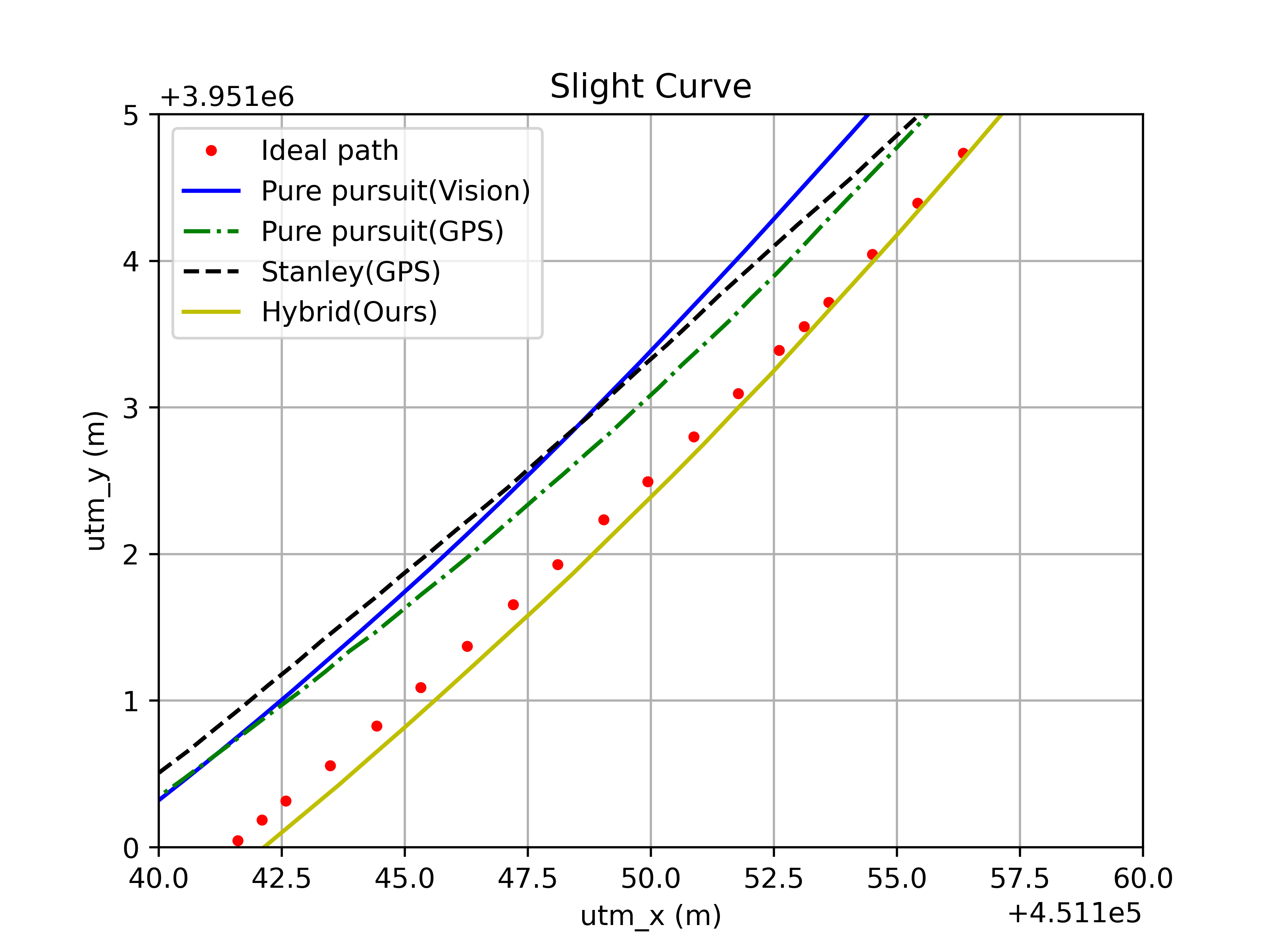}}
\subfloat[ ]{\includegraphics[width=3.5cm, height =3.5cm]{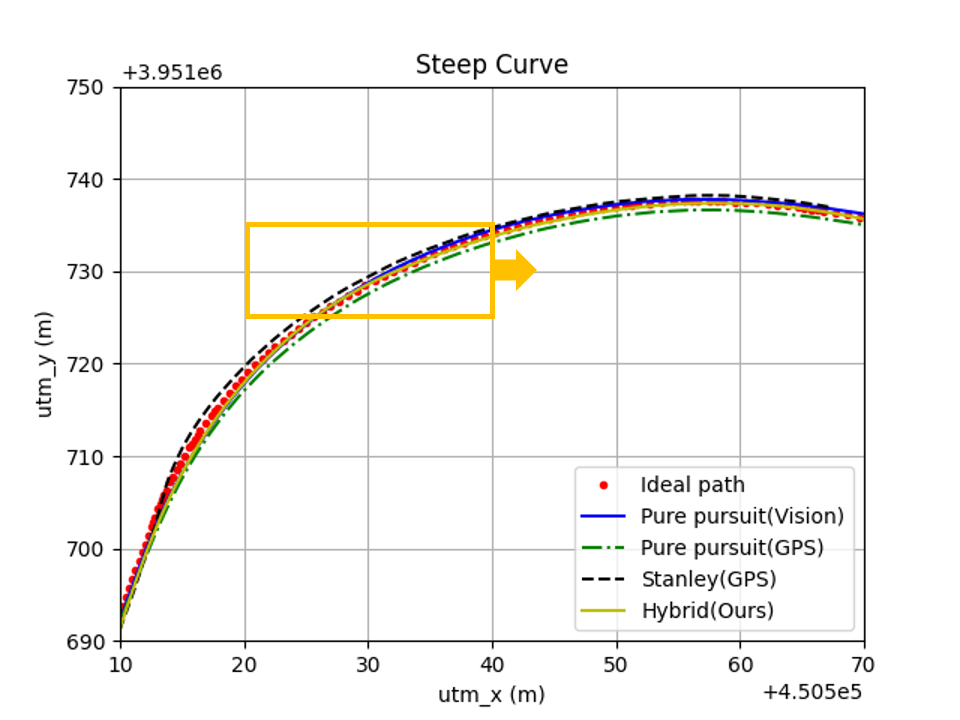}}
\subfloat[ ]{\includegraphics[width=3.5cm, height =3.5cm]{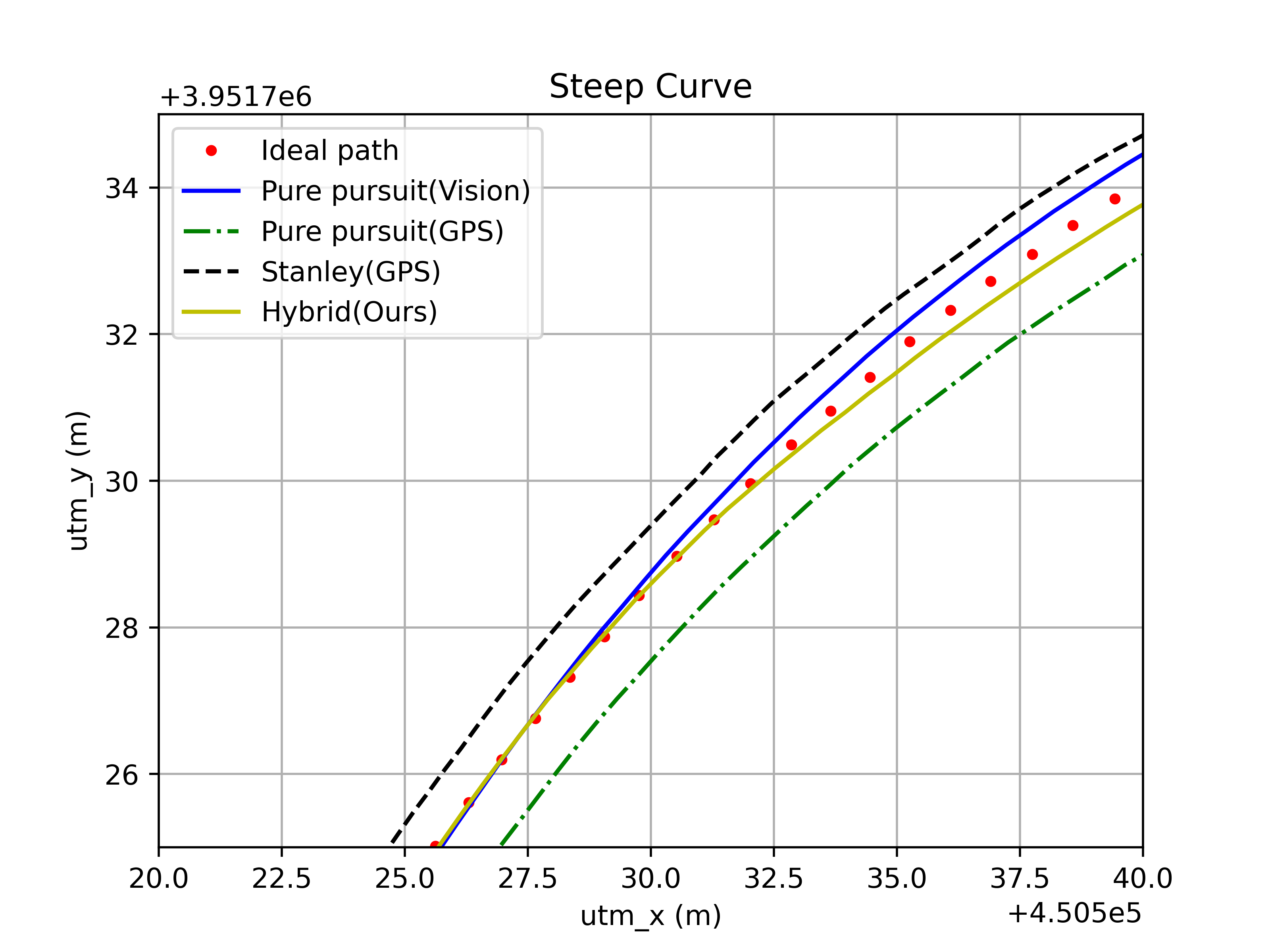}}
\caption{Test result at normal road environments (a) straight road, (b) slight curve, (c) enlarged graph of (b), (d) steep curve, (e) enlarged graph of (d).}
\label{fig: test result normal}
\end{figure*}

\begin{table*}[hbt]
    \begin{center}
    \renewcommand{\arraystretch}{1.1}
        \caption{Quantitative results by different path tracking algorithms under normal road environments (\ie, straight, slight curve, steep curve).}
        \label{tab: normal}
        \addtolength{\tabcolsep}{2pt}
        \resizebox{\textwidth}{!}{%
        \begin{tabular}{c|c|c|ccc|cc}
        \multirow{2}{*}{Scenes} & \multirow{2}{*}{Methods} & \multirow{2}{*}{$Success$} & \multicolumn{3}{c|}{Tracking Performance} &  \multicolumn{2}{c}{Driving Stability} \\ \cline{4-8}
        & & &RMSE ($lateral$) $\downarrow$ & RMSE ($longitudinal$) $\downarrow$ & $Distance$ $\downarrow$ & RMSE ($yaw$) $\downarrow$ & RMSE ($steer$) $\downarrow$ \\
        \Xhline{2.5\arrayrulewidth}
        \multirow{4}{*}{Straight} & Pure pursuit (vision) & $\bigcirc$ &  \textbf{0.091} & \textbf{0.019} & \textbf{0.068} & \textbf{0.079} & \textbf{1.314}\\
        \cline{2-8} &Pure pursuit (GPS) & $\bigcirc$ & 0.223 & \textbf{0.007} & 0.155 & 0.174 & 1.755\\
        \cline{2-8} &Stanley (GPS) & $\bigcirc$ & \textbf{0.196} & 0.065 & \textbf{0.146} & 0.158 & 2.406\\
        \cline{2-8} &Hybrid & $\bigcirc$ & 0.377 & 0.049 & 0.285 & \textbf{0.025} & \textbf{1.300}\\
        \Xhline{1.5\arrayrulewidth}
        \multirow{4}{*}{ Slight curve } & Pure pursuit (vision) & $\bigcirc$ & \textbf{0.027} & \textbf{0.283} & \textbf{0.207} & 17.742 & \textbf{2.270} \\
        \cline{2-8} &Pure pursuit (GPS) & $\bigcirc$ & 0.239 & 0.577 & 0.598 & \textbf{16.685} & 2.894\\
        \cline{2-8} &Stanley (GPS) & $\bigcirc$ & 0.131 & 0.553 & 0.463 & \textbf{16.247} & 3.430\\
        \cline{2-8} &Hybrid & $\bigcirc$ & \textbf{0.027} & \textbf{0.169} & \textbf{0.144} & 16.933 & \textbf{1.929}\\
        \Xhline{1.5\arrayrulewidth}
        \multirow{4}{*}{Steep curve} & Pure pursuit (vision) & $\bigcirc$ & 0.450 & 0.397 & 0.522 & \textbf{57.994} & \textbf{2.865} \\
        \cline{2-8} &Pure pursuit (GPS) & $\bigcirc$ & \textbf{0.129} & 0.492 & \textbf{0.384} & 62.959 & \textbf{3.199}\\
        \cline{2-8} &Stanley (GPS) & $\bigcirc$ & \textbf{0.145} & \textbf{0.174} & \textbf{0.166} & 63.501 & 5.406\\
        \cline{2-8} &Hybrid & $\bigcirc$ & 0.321 & \textbf{0.370} & 0.414 & \textbf{52.047} & 3.485\\
        \hline
        \end{tabular}%
        }
        \begin{tablenotes}
        \small
        \item The unit of RMSE ($lateral$), RMSE ($longitudinal$) and $Distance$ is meter. The unit of RMSE ($yaw$) and RMSE ($steer$) is degree.
        \end{tablenotes}
    \end{center}
\end{table*}

Fig. \ref{fig: test result normal} shows graphically the tracking results of Pure pursuit (vision), Pure pursuit (GPS), Stanley (GPS) and Hybrid in three different scenes.
In Fig. \ref{fig: test result normal}(a), (b) and (d) shows the results; a straight road, a slight curve road, and a steep curve road. Graph (c) and (e) are enlarged views of the yellow boxes in graphs (b) and (d) in Fig. \ref{fig: test result normal}, respectively. It can be seen in the figure that Pure pursuit (vision), Pure pursuit (GPS), Stanley (GPS), and Hybrid are all almost identical to the ideal path in the normal road environment. Numerical interpretation of the graph is demonstrated in TABLE \ref{tab: normal}.

Thus, it can be concluded that all of modified trackers successfully drove the suggested section without infringement on the lane from the $Success$ column in TABLE \ref{tab: normal}). As mentioned in section \ref{subsection: path tracking algorithm} in this paper, each tracker clearly shows high performance in the certain section of the road due to the characteristics of each tracker. Therefore, this implies that the result of modified each tracker produces optimal performance for each situation. Hybrid system shows intermediate or higher results in the overall result. Since the Hybrid system operates by selecting several optimal trackers at a fast computational speed, it is possible to secure at least the average performance of each tracker.

In the case of straight road, the overall tracking performance (\ie, RMSE) of Hybrid system is lower than the peak performance of other base trackers. 
It is because Hybrid is the system composed of the above three trackers. The Hybrid system also guarantees of both stability and tracking performance, whereas these slight losses may occur. As shown in TABLE \ref{tab: normal} and \ref{tab: complex}, in terms of stability, the Hybrid achieves the highest performance. RMSE ($yaw$) and RMSE ($steer$) values were lower than the second best performances of Pure pursuit (vision) by 0.054 and 0.014 in degree unit respectively. Thus, it convinces that Hybrid is superior in terms of driving stability. It is also found that Hybrid has a tendency to lack the ability to compensate for residual errors in a straight road situation as it is changed to multiple trackers due to the fast selection speed. However, due to these characteristics, robustness of the stability for the Hybrid system improved the overall driving performance.

The driving performances show the most effective in the slight curve due to the characteristics of Hybrid. 
It is also able to be found that other roads have a clear difference in performance between vision and GPS, but slight curve shows relatively similar performance. In other words, the Hybrid has the best overall performance in slight curve because the tracking performance of multiple trackers is properly used. 
Conversely, RMSE ($yaw$) leads to 16.933 degrees out of ideal path, which is the third best among the four algorithms. However, the difference between the value of the best algorithm and Hybrid was only 0.686 degrees which equals to only 0.04\% differences with the results of the other algorithms. Since the value of RMSE ($steer$) was the smallest one, it can be explained that the driving stability of Hybrid system provides the best comfortableness.

Moreover, in the steep curve, due to the limitations of vision, trackers using GPS are mainly used. In this case, Stanley (GPS) shows high precision for this road environment. However, due to bad stability, Hybrid uses only Pure pursuit (vision) and Pure pursuit (GPS) to secure overall driving performance. As an experimental result, the tendency of error becomes similar that of Pure pursuit (vision). The tracking performance of Hybrid gets worse than the performance of Stanley (GPS) since Hybrid uses Pure pursuit trackers. Thus, it can be deduced that relatively low tracking performance of Hybrid system is a result of the algorithm operation to ensure stability. When human drives on the steep curve, they usually do offset driving for driving stability. Since the proposed system does not have environment recognition, it is important to precisely follow the provided path. Hence, when the importance of the in-track is high, such as the steep curve, the Hybrid drives by focusing on the in-track rather than the stability. In the steep curve, the results show Stanley (GPS) has the best tracking performance, but show Stanley (GPS) has poor driving stability. Therefore, it can be concluded that Hybrid secured driving stability by properly switching two trackers, Pure pursuit (vision) and Pure pursuit (GPS), which have good stability performance in the secured tracking performance situation. Considering the value of RMSE ($steer$) compared to humans with offset driving, the smaller value of RMSE ($yaw$) than 5 degrees compared to other trackers imply that the overall driving performance of Hybrid in steep curve represents superior.

\begin{figure*}[hbt]
\centering
\subfloat[ ]{\includegraphics[width=8cm, height =7cm]{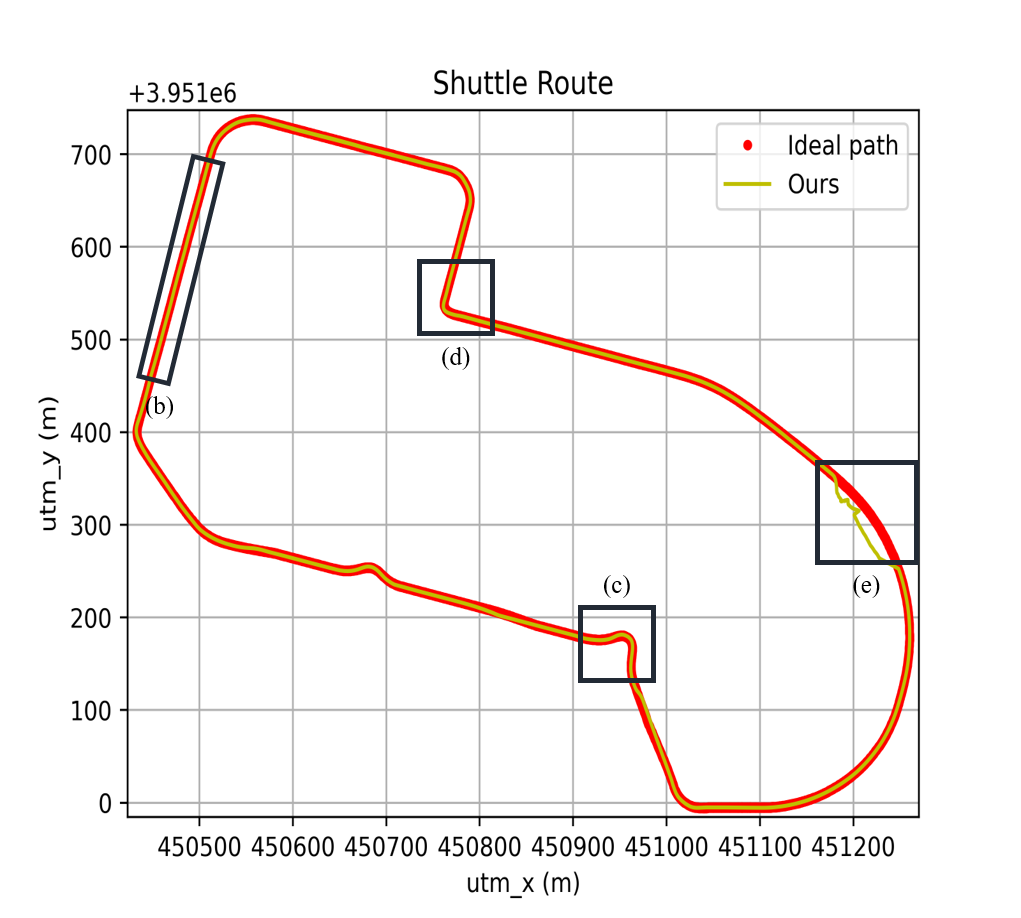}}\hfill
\subfloat[ ]{\includegraphics[width=8cm, height =7cm]{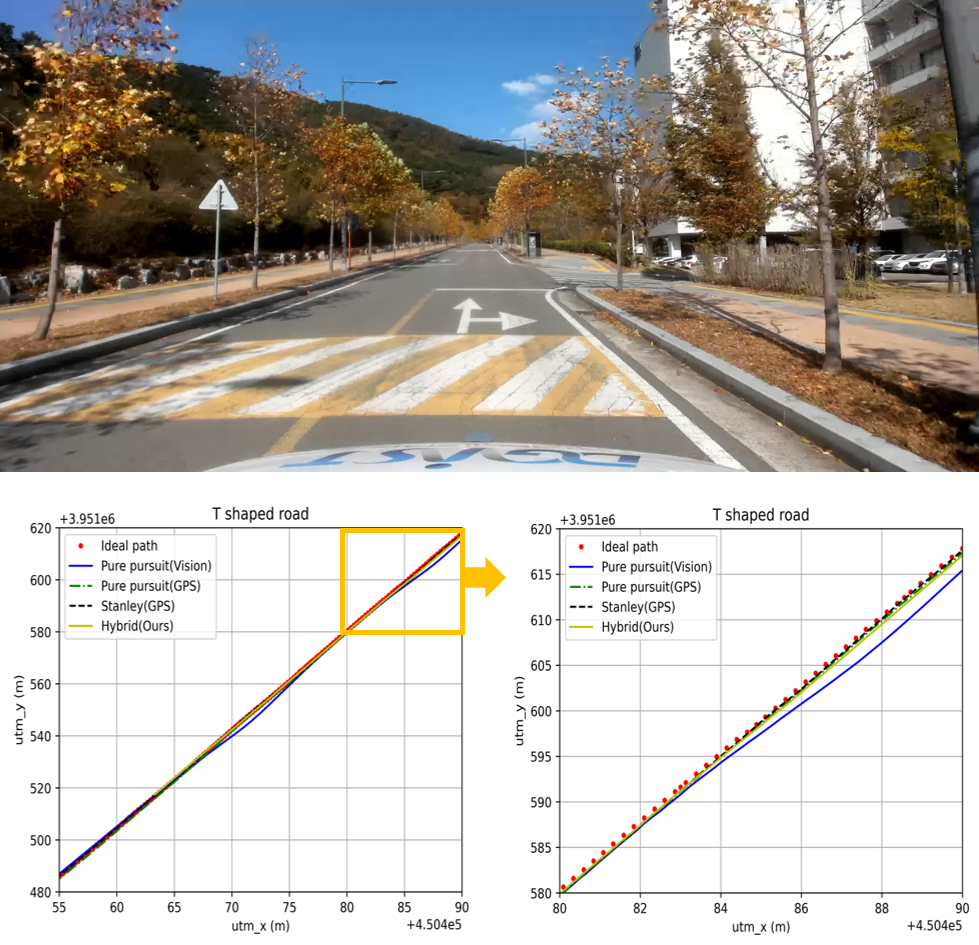}}\hfill
\subfloat[ ]{\includegraphics[width=8cm, height =7cm]{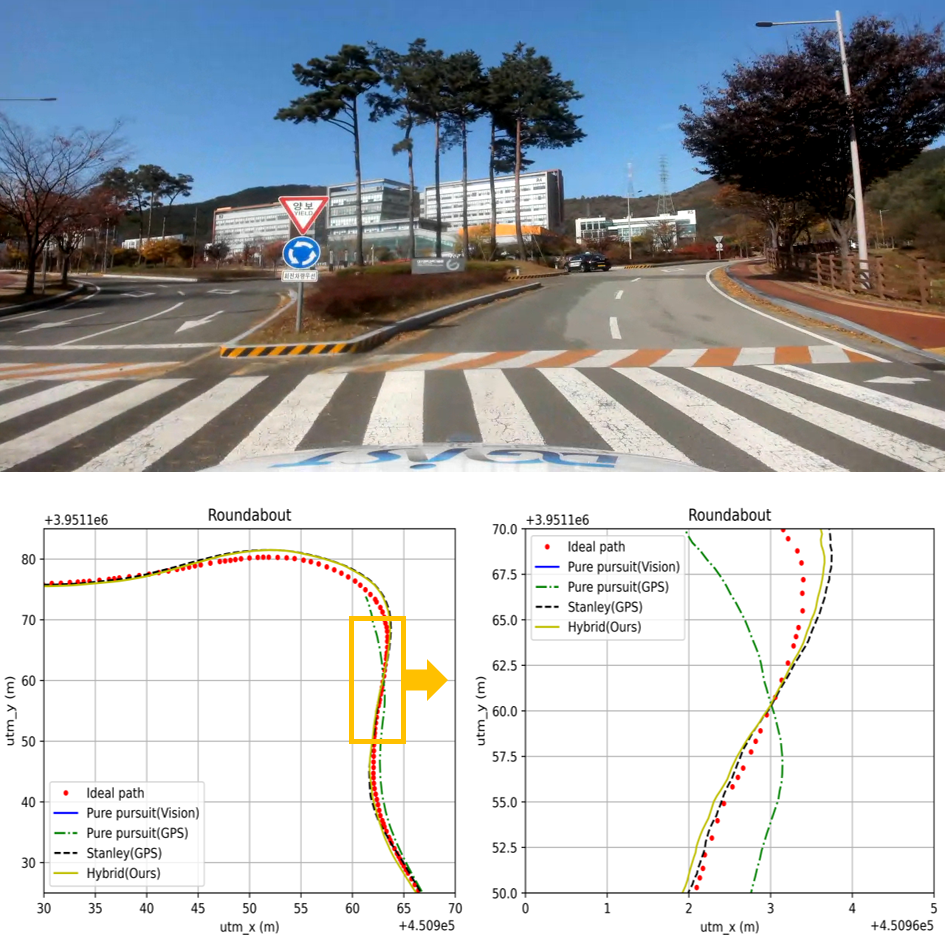}}\hfill
\subfloat[ ]{\includegraphics[width=8cm, height =7cm]{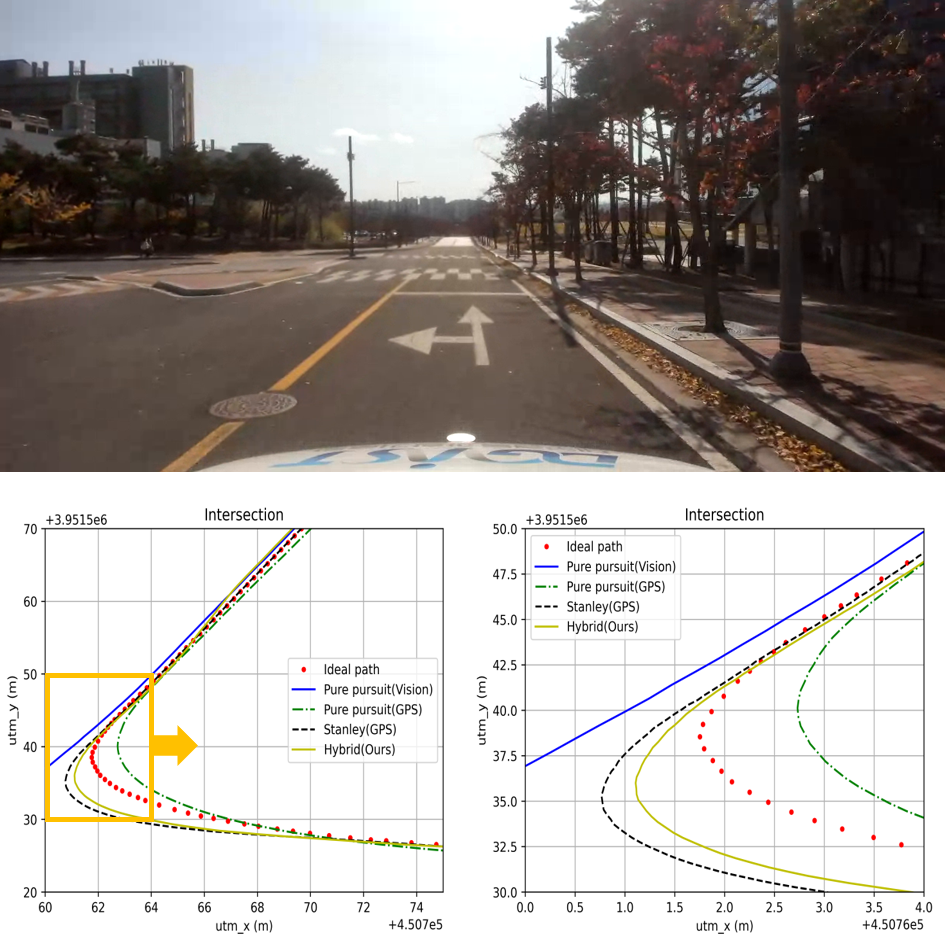}}\hfill
\caption{DGIST shuttle route and experimental results at complex road environments (a) DGIST shuttle route, (b) straight road with multiple 3-way junctions, (c) roundabout, (d) intersection.}
\label{fig: shuttle route, complex road environment}
\end{figure*}

\begin{table*}[hbt]
    \begin{center}
        \renewcommand{\arraystretch}{1.1}
        \caption{Quantitative results by different path tracking algorithms under complex road environments (\ie, straight road with multiple 3-way junctions, roundabout, tunnel, intersection).}
        \label{tab: complex}
        \addtolength{\tabcolsep}{2pt}
        \resizebox{\textwidth}{!}{%
        \begin{tabular}{c|c|c|ccc|cc}
        \multirow{2}{*}{Scenes} & \multirow{2}{*}{Methods} & \multirow{2}{*}{$Success$} & \multicolumn{3}{c|}{Tracking Performance} &  \multicolumn{2}{c}{Driving Stability} \\ \cline{4-8}
        & & &RMSE ($lateral$) $\downarrow$ & RMSE ($longitudinal$) $\downarrow$ & $Distance$ $\downarrow$ & RMSE ($yaw$) $\downarrow$ & RMSE ($steer$) $\downarrow$ \\
        \Xhline{2.5\arrayrulewidth}
        \multirow{4}{*}{\begin{tabular}[c]{@{}c@{}}Straight road\\ with multiple\\ 3-way junctions\end{tabular}} & Pure pursuit (vision) & $\bigcirc$ & 0.152 & 1.063 & 0.862 & 0.461 & \textbf{1.671} \\
        \cline{2-8} &Pure pursuit (GPS) & $\bigcirc$ & \textbf{0.029} & 0.290 & 0.219 & 2.871 & 2.434\\
        \cline{2-8} &Stanley (GPS) & $\bigcirc$ & 0.081 & \textbf{0.183} & \textbf{0.142} & \textbf{0.125} & 2.734\\
        \cline{2-8} &Hybrid & $\bigcirc$ & \textbf{0.056} & \textbf{0.214} & \textbf{0.157} & \textbf{0.018} & \textbf{1.660}\\
        \Xhline{1.5\arrayrulewidth}
        \multirow{4}{*}{Roundabout} & Pure pursuit (vision) & \ding{53} & - & - & - & - & 8.252 \\
        \cline{2-8} &Pure pursuit (GPS) & \ding{53} & - & - & - & - & 6.431 \\
        \cline{2-8} &Stanley (GPS) & $\bigcirc$ & 0.415 & \textbf{0.403} & 0.471 & 187.166 & \textbf{5.912}\\
        \cline{2-8} &Hybrid & $\bigcirc$ & \textbf{0.232} & 0.486 & \textbf{0.377} & \textbf{172.815} & \textbf{5.708}\\
        \Xhline{1.5\arrayrulewidth}
        \multirow{4}{*}{Intersection} & Pure pursuit (vision) & \ding{53} & - & - & - & - & 6.664\\
        \cline{2-8} &Pure pursuit (GPS) & $\bigcirc$ & \textbf{0.130} & 0.263 & \textbf{0.227} & 75.829 & 6.046\\
        \cline{2-8} &Stanley (GPS) & $\bigcirc$ & \textbf{0.424} & \textbf{0.109} & \textbf{0.260} & \textbf{73.313} & \textbf{5.493}\\
        \cline{2-8} &Hybrid & $\bigcirc$ & 0.439& \textbf{0.181} & 0.330 & \textbf{65.093} & \textbf{5.481}\\
        \Xhline{1.5\arrayrulewidth}
        \multirow{4}{*}{Tunnel} & Pure pursuit (vision) & $\bigcirc$ &  - & - & - & - & \textbf{1.935} \\
        \cline{2-8} &Pure pursuit (GPS) & \ding{53} & - & - & - & - & 7.704\\
        \cline{2-8} &Stanley (GPS) & \ding{53} & - & - & - & - & 8.305\\
        \cline{2-8} &Hybrid & $\bigcirc$ & - & - & - & - & \textbf{1.911}\\
        \hline
        \end{tabular}%
        }
        \begin{center}
        \begin{tablenotes}
        \small
        \item The unit of RMSE ($lateral$), RMSE ($longitudinal$) and $Distance$ is meter. The unit of RMSE ($yaw$) and RMSE ($steer$) is degree.
        \end{tablenotes}
        \end{center}
    \end{center}
\end{table*}

\subsubsection{Test Results Under Complex Road Environment \label{subsubsection: test results under complex road environment}}
In the previous section, it was found that three tracking algorithms and Hybrid system succeeded in driving the normal road environment. However, there were some sections where three algorithms except Hybrid could not properly control the vehicle in roundabout, intersection, and tunnel shown in the Fig. \ref{fig: shuttle route, complex road environment}. 
The Fig. \ref{fig: shuttle route, complex road environment}(a) shows that the path in which the proposed system actually drove the entire DGIST shuttle route closely matches the ideal path. Each road in the complex road environment is marked by a black box in Fig. \ref{fig: shuttle route, complex road environment}(a). Figures of (b), (c) and (d) comprise a combination of the roadway actual tested, the driving graph and the graph extending the yellow box. Fig. \ref{fig: shuttle route, complex road environment}(b) shows the results of tracking straight roads with multiple 3-way junctions and the enlarged graph demonstrates that Pure pursuit (vision) has poor tracking performance due to the absence of lanes in the area with 3-way junctions. Fig. \ref{fig: shuttle route, complex road environment}(c) displays the results of tracking the roundabout. Pure pursuit (vision) was unable to follow because the lane of the roundabout was not clear and Pure pursuit (GPS) was unable to follow accurately at very large curvature. Only Stanley controller succeeded in tracking and the proposed system tracked well by selecting Stanley controller among the three tracking algorithms. Fig. \ref{fig: shuttle route, complex road environment}(d) shows the results of tracking the intersection. Pure pursuit (vision) failed to follow the driving guidance line of the intersection. Pure pursuit (GPS) and Stanley controller tracked the path well. Pure pursuit (GPS) adjusted the steers quickly and Stanley controller adjusted the steers slowly, whereas Pure pursuit (GPS) occasionally crossed the center line and tracked the path overall.

In the complex road environment, a numerical analysis of the tracking performance and driving stability of each algorithm are listed in TABLE \ref{tab: complex}. 
In straight road with multiple 3-way junctions, the algorithms using the GPS based path have better tracking performance than the algorithm using the vision based path since lanes are often disconnected. The $distance$ value of Hybrid is 0.157 meter, which can be seen to have a value between the $distance$ value of Stanley (GPS) and the $distance$ value of Pure pursuit (GPS). It might be implied that the Hybrid selected alternately Stanley (GPS) and Pure pursuit (GPS). In particular, the values of RMSE ($yaw$) and RMSE ($steer$) representing a driving stability showed the best among the tested methods.

In roundabout, due to the few lanes, Pure pursuit (vision) was unable to drive properly this road. Also, it was unable to adopt Pure pursuit (GPS) due to very large curvature. The TABLE \ref{tab: complex} also indicated that Stanley (GPS) and Hybrid succeeded in driving the roundabout, and $distance$ value of Hybrid is about 0.094 meter closer than that of Stanley (GPS)'s $distance$ to the ideal path.
In terms of driving stability, RMSE ($yaw$) value of Hybrid showed 14.351 degrees better than that of Stanley (GPS), and RMSE ($steer$) of Hybrid is slight difference with that of Stanley (GPS). This is the reason why there is a big difference (\ie, 14.3 degrees) in RMSE ($yaw$) even though Hybrid adopted Stanley (GPS) because Hybrid chose different trackers in entering the roundabout. It can be finally induced that Hybrid becomes superior in terms of tracking performance and driving stability.

In intersection, Pure pursuit (GPS), Stanley (GPS), Hybrid system succeeded. The $distance$ values show all 0.330 meter or less, so the tracking performance indicates high. Although Hybrid did not represent the highest tracking performance among them, it guarantees the driving stability. 

\begin{figure}[hbt]
\centering
\subfloat[ ]{\includegraphics[width=8cm, height =4cm]{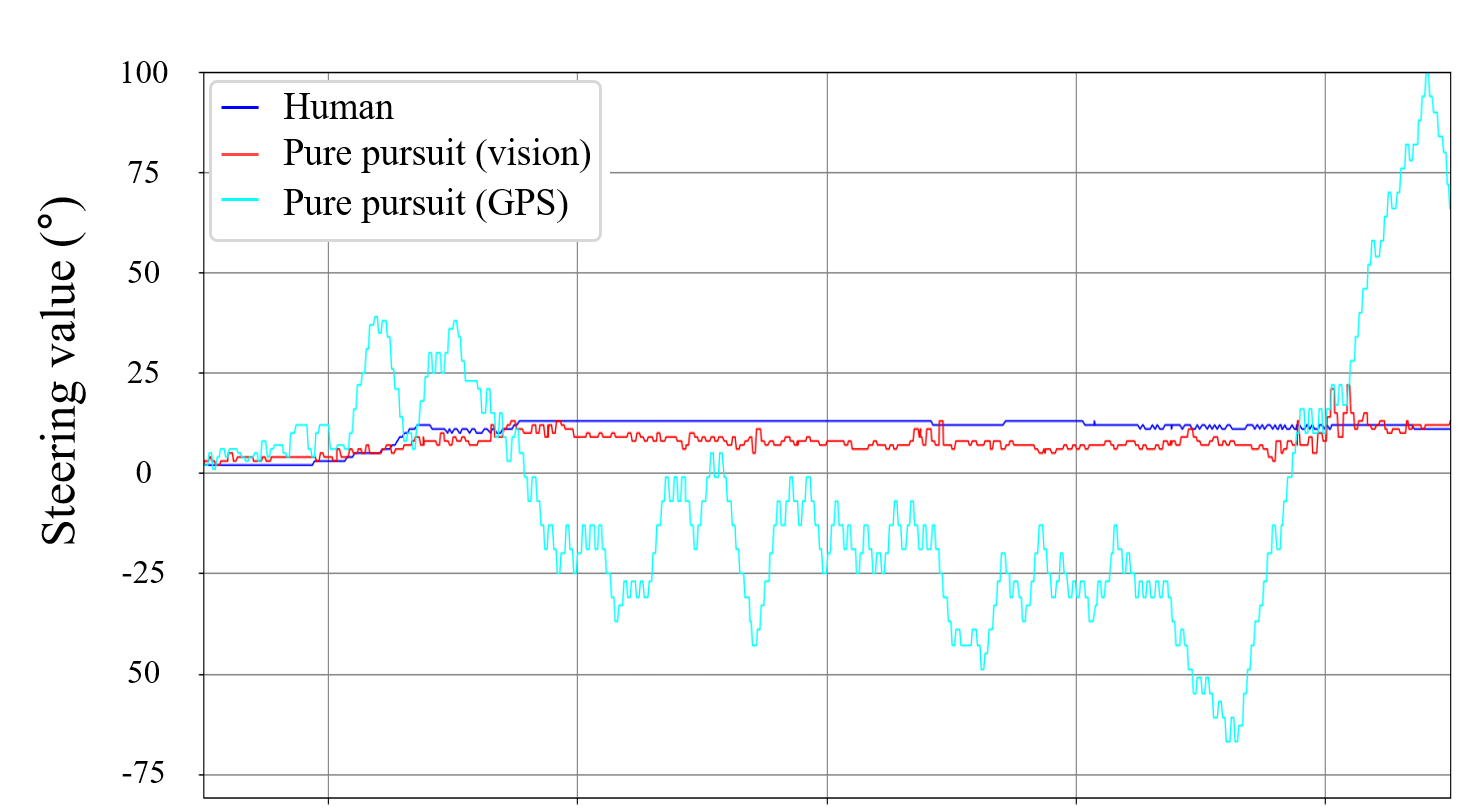}}\hfill
\subfloat[ ]{\includegraphics[width=8cm, height = 4 cm]{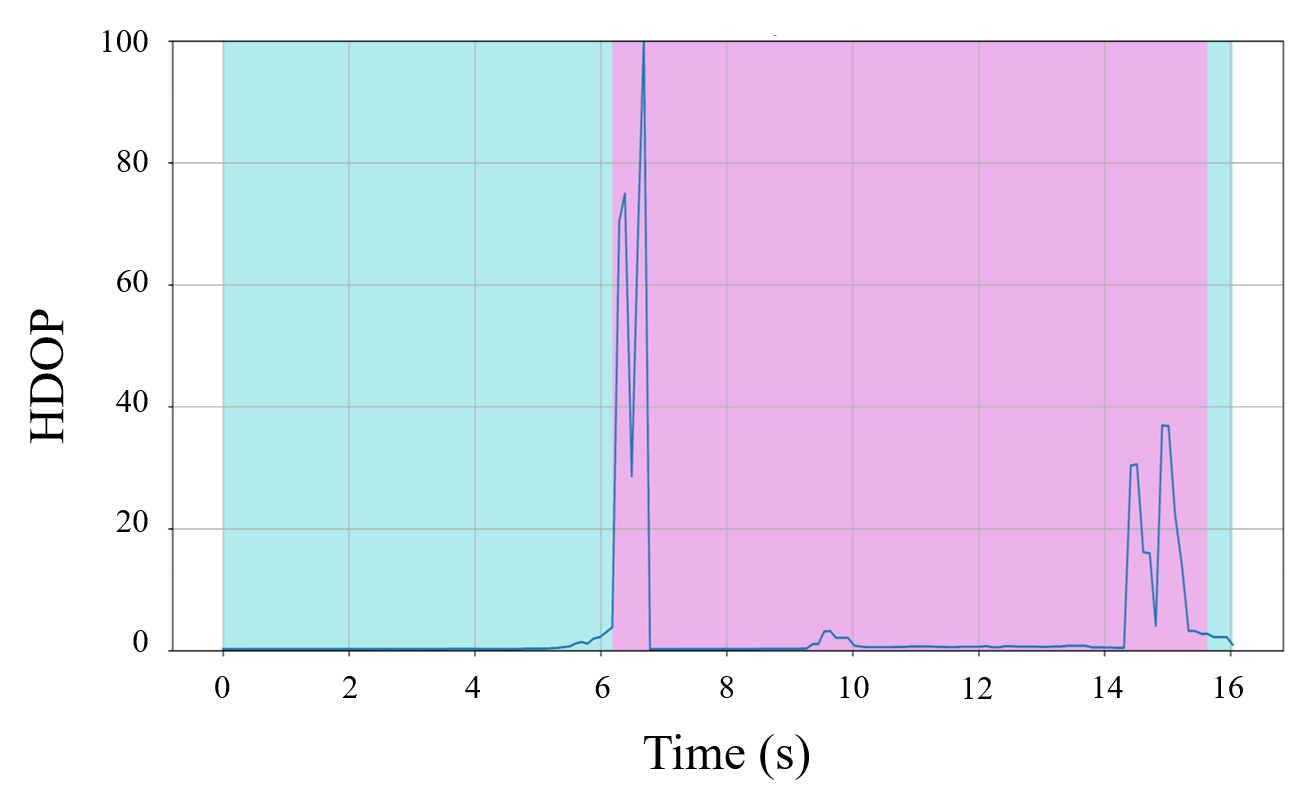}}\hfill
\caption{Experimental results in tunnel (a) comparison of steering angle value, (b) GPS reliability.}
\label{fig: steer graph in tunnel}
\end{figure}

In Tunnel, the GPS did not work properly, so tracking performance could not be obtained. The tracking of the tunnel area was operated well based on Pure pursuit (vision), but the reliability of the GPS showed low and the graph seems that the vehicle did not follow the path, as in the (e) area of Fig. \ref{fig: shuttle route, complex road environment}(a). It is shown that the steer value driven by humans and the steer values predicted by each algorithm. Fig. \ref{fig: steer graph in tunnel}(b) shows reliability of GPS. When the reliability of GPS becomes high, the graph background color is highlighted in cyan, otherwise in magenta. 
The steer value provided by Pure pursuit (vision) shows pretty similar to the steer value driven by humans, whereas the steer value provided by Pure pursuit (GPS) does not. It is also shown in Fig. \ref{fig: steer graph in tunnel}(a) that algorithms using GPS based path are not able to pass through tunnels due to the low reliability of GPS. 

Finally, since GPS based path was impossible to operate correctly in the tunnel, only RMSE (steer) was obtained. RMSE ($steer$) of Pure pursuit (vision) and RMSE ($steer$) of Hybrid are similar with 0.6 degrees differences. 
Therefore, Hybrid succeeded in passing through the tunnel using Pure pursuit (vision).
In summary of experimental results, the path tracking algorithms for the vision based path and the GPS based path were appropriately switched through the Hybrid tracker based optimal path tracking system to drive the shuttle route on DGIST campus.
\section{Conclusion}\label{section: conclusion}
This paper proposes Hybrid tracker based optimal path tracking system for autonomous vehicles. The designed system combines, evaluates and applies several trackers to driving on complex road conditions. The proposed system includes all of the processing algorithms of deep learning based lane detection algorithms, coordinate system transformation, three modified geometric trackers and optimal path selection algorithm. Based on combining all above processes, 
the proposed system guarantees reasonable driving performance for trade-off of driving stability and tracking performance in both normal and complex road environments. This study also notes that the quantitative comparison using ideal path and the steering value from human shows the overall driving performance of Hybrid tracker based optimal path tracking system. With experimental results, this study convinces to build a seamless system from the sensor part of the autonomous vehicle to the action part for better performance. In other words, the key to build this system was that control and sensor recognition should be made in one process. Consequently, as shown in experimental results, it notes that proposed system of the optimal path selection and system architecture significantly improves both driving stability and tracking performance in the presence of high complexity of road conditions.

\bibliographystyle{IEEEtran}
\bibliography{access.bib}

\begin{IEEEbiography}[{\includegraphics[width=1in,height=1.25in,clip,keepaspectratio]{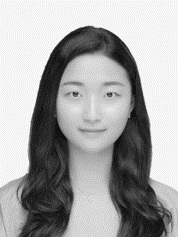}}]{EUNBIN SEO} was born in Suwon, South Korea, in 1999. She is currently pursuing the B.S. degree in computer science and technology with Daegu Gyeongbuk Institute of Science and Technology (DGIST), Daegu, Korea. Her research interests include computer vision and automatic control.
\end{IEEEbiography}

\begin{IEEEbiography}[{\includegraphics[width=1in,height=1.25in,clip,keepaspectratio]{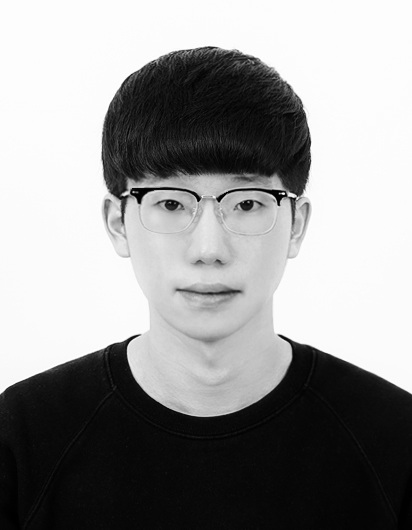}}]{SEUNGGI LEE} was born in Seoul, South Korea, in January 2000. He is currently pursuing the B.S. degree  in electrical engineering from Daegu Gyeongbuk Institute of Science and Technology (DGIST), Daegu, Korea. His research interest include communication system and wireless power transfer.
\end{IEEEbiography}

\begin{IEEEbiography}[{\includegraphics[width=1in,height=1.25in,clip,keepaspectratio]{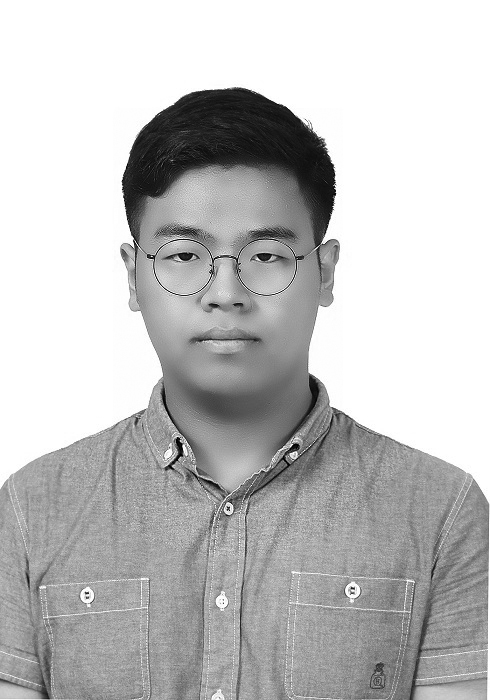}}]{GWANJUN SHIN} was born in Seosan, South Korea in 1999. He is currently undergraduate students at College of Trans-disciplinary studies, Daegu Gyeongbuk Institute of Science and Technology (DGIST). His research interests include visual recognition and domain adaptation in machine learning. He is also the NVIDIA Deep learning institute University Ambassadorship and instructor.
\end{IEEEbiography}

\begin{IEEEbiography}[{\includegraphics[width=1in,height=1.25in,clip,keepaspectratio]{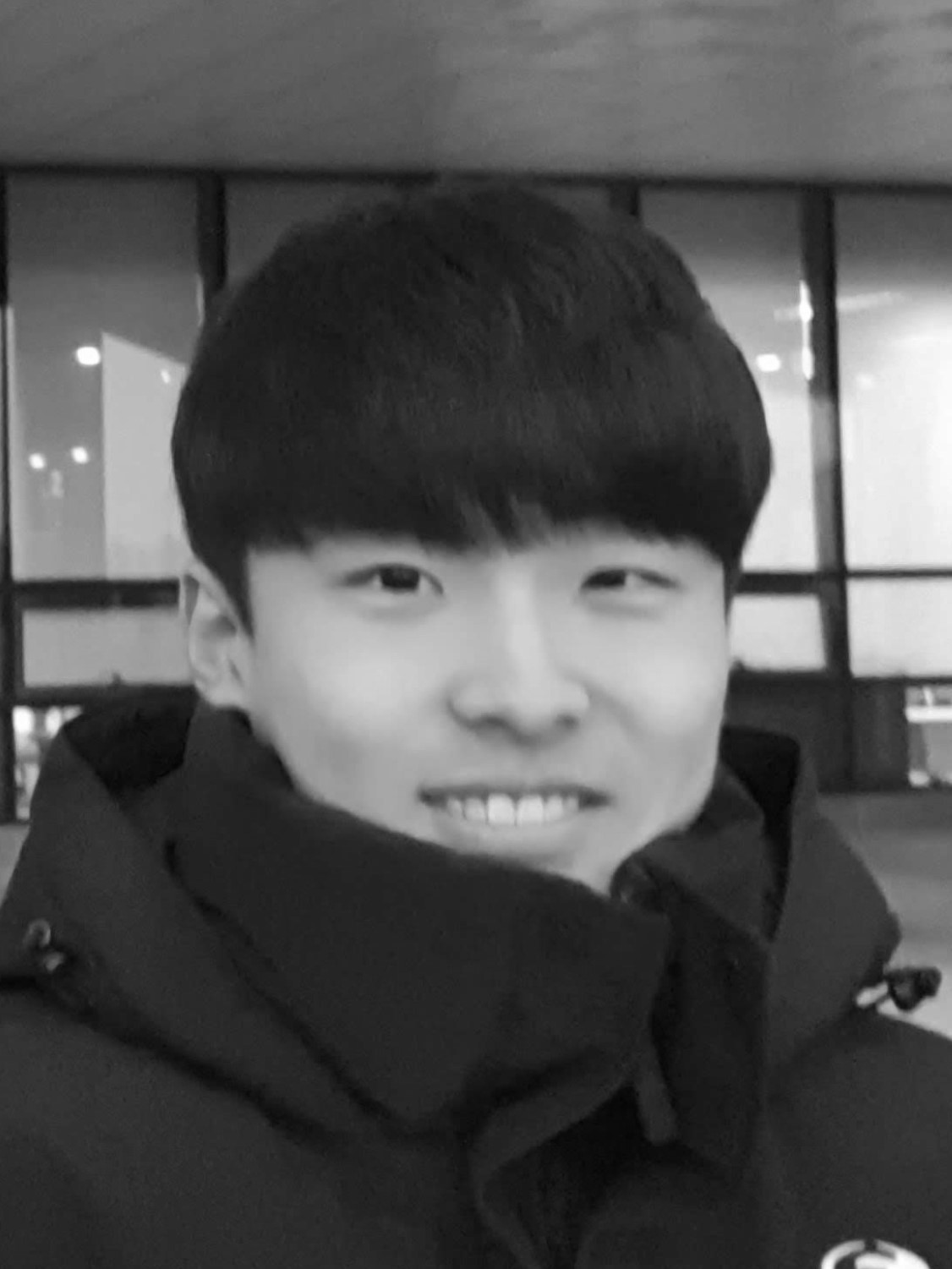}}]{HOYEONG YEO} was born in Daegu, South Korea in 1999, He is currently undergraduate students at College of Trans-disciplinary studies, Daegu Gyeongbuk Institute of Science and Technology (DGIST). His research interests include motion control and deep learning.
\end{IEEEbiography}

\begin{IEEEbiography}[{\includegraphics[width=1in,height=1.25in,clip,keepaspectratio]{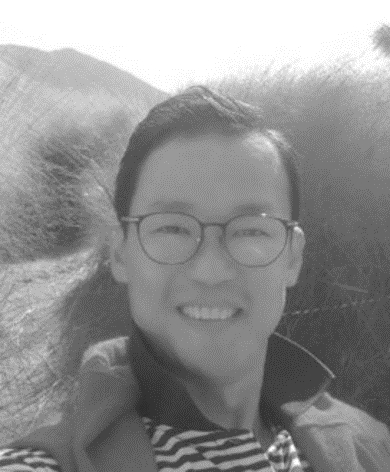}}]{YONGSEOB LIM} received the Ph.D. degree in mechanical engineering from the University of Michigan, Ann Arbor, MI, USA, in 2010. He was with the Hyundai Motor Company R\&D Advanced Chassis Platform Research Group, South Korea, as a Research Engineer. His works at Hyundai Motor Company focused on advanced vehicle dynamics and control technologies, including active suspension, steering and braking control systems. Moreover, He also with Samsung Techwin Company Mechatronics System Development Group, South Korea as a Principal Research Engineer. His works at Samsung focused on developing stabilization control algorithm for the Remotely Controlled Robotic Arms. He is currently an Associate Professor in the Robotics Engineering Department, Daegu Gyeongbuk Institute of Science and Technology (DGIST), and the Co-Director of the Joint Research Laboratory of Autonomous Systems and Control (ASC) and Vehicle in the Loop Simulation (VILS) Laboratories. His research interests include the modeling, control, and design of mechatronics systems with particular interests in autonomous ground vehicles and flight robotics systems and control.
\end{IEEEbiography}

\begin{IEEEbiography}[{\includegraphics[width=1in,height=1.25in,clip,keepaspectratio]{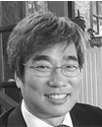}}]{GYEUNGHO CHOI} received the Ph.D. degree in mechanical engineering from the University of Alabama, AL, USA, in 1992, also Honorary PhD Degree in Automotive \& Energy Engineering conferred be Her Royal Highness Princess Maha Chakri Sirindhorn from King Mongkut’s University of Technology North Bangkok, Thailand in 2018. He served President in Korean Auto-vehicle Safety Association and Editor-in-Chief in Korean Society for Engineering Education. He has taught Thermodynamics, Heat Transfer, and Automotive Engineering in Keimyung University and Daegu Gyeongbuk Institute of Science and Technology as a Professor for 28 years. He is also the Co-Director of the Joint Research Laboratory of Autonomous Systems and Control (ASC) and Vehicle-in-the-Loop Simulation (VILS) Laboratories. His research interests include self-driving vehicle, ADAS system, Vehicle-In-the-Loop system, and design of alternative energy utilization.
\end{IEEEbiography}
\EOD
\end{document}